%% file: iclr2027_conference.tex
\documentclass{article}
\usepackage{iclr2027_conference,times}

\input{math_commands.tex}

\usepackage{xcolor}\definecolor{customcitecolor}{HTML}{3E7FA3}\usepackage[colorlinks=true,citecolor=customcitecolor,linkcolor=red,urlcolor=magenta]{hyperref}
\usepackage{url}
\usepackage{graphicx}

\usepackage{amssymb}

\usepackage{wrapfig}
\usepackage{algpseudocode}

\usepackage{booktabs}
\usepackage{multirow}

\title{Connected Self Forcing: Beyond Local Learning in Video Autoregression}
\author{\begin{minipage}[t]{\dimexpr\textwidth-2\tabcolsep\relax}\raggedright
\bfseries\small
Dongbin Zhang\textsuperscript{1}\hfill Chaoda Zheng\textsuperscript{1}\thanks{Project lead.\qquad$^\dagger$Corresponding author.}\hfill Kangjie Chen\textsuperscript{1}\hfill Xiangyu Li\textsuperscript{1}\hfill Shijia Chen\textsuperscript{1}\hfill Jinhao Deng\textsuperscript{1}\\
Yuqi Zhang\textsuperscript{1}\hfill Guangfeng Jiang\textsuperscript{1}\hfill Hongbin Lin\textsuperscript{2}\hfill Choo Sin Wai\textsuperscript{3}\hfill Minqi Wang\textsuperscript{2}\hfill Puyi Wang\textsuperscript{2}\\
Jingye Zhang\textsuperscript{3}\quad Yu Zhang\textsuperscript{1}\quad Xianming Liu\textsuperscript{1}\quad Boyang Wang\textsuperscript{1$\dagger$}\\[0.8ex]
\normalfont\footnotesize
\textsuperscript{1}XPeng\quad\textsuperscript{2}The Chinese University of Hong Kong\quad
\textsuperscript{3}Tsinghua University
\end{minipage}}

\iclrfinalcopy
\begin{document}

\maketitle
\lhead{Preprint}

\input{sec/0_abstract.tex}

\input{sec/1_intro.tex}

\input{sec/2_related.tex}

\input{sec/3_method.tex}

\input{sec/4_experiment.tex}

\input{sec/5_conclusion.tex}

\bibliography{iclr2027_conference}
\bibliographystyle{iclr2027_conference}

\input{sec/7_appendix.tex}

\end{document}

%% file: math_commands.tex
\usepackage{amsmath,amsfonts,bm}

\def\eqref#1{equation~\ref{#1}}

\def\1{\bm{1}}

\DeclareMathAlphabet{\mathsfit}{\encodingdefault}{\sfdefault}{m}{sl}
\SetMathAlphabet{\mathsfit}{bold}{\encodingdefault}{\sfdefault}{bx}{n}



%% file: sec/0_abstract.tex
\begin{figure}[ht]
  \centering
  \includegraphics[width=\textwidth]{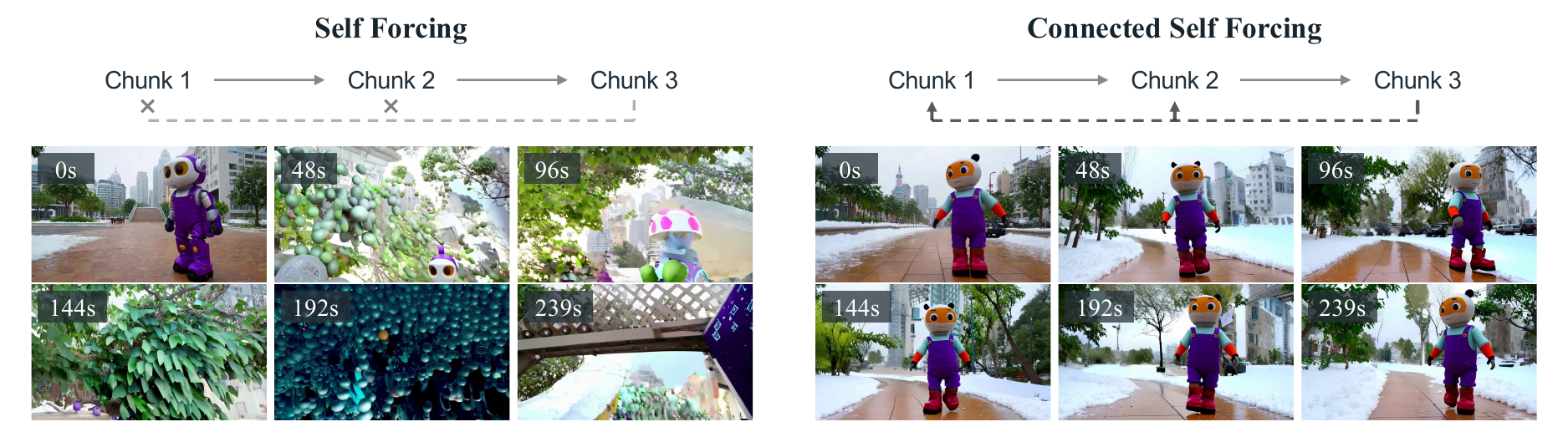}
  \caption{Connected Self Forcing restores gradient flow from later chunks to
    earlier ones, helping maintain visual consistency in long videos.}
  \label{fig:teaser}
  \vspace{-10pt}
\end{figure}

\begin{abstract}

To stream long videos while maintaining visual quality and temporal consistency, Self Forcing mitigates exposure bias through self-rollout training on self-generated histories with key-value (KV) caching. To keep memory manageable, it detaches historical caches, preserving forward dependencies between chunks but severing the backward gradient paths.
We introduce Connected Self Forcing, a training framework that reconnects gradient paths across autoregressive chunks, allowing feedback from later predictions to guide how earlier context is generated. These connections go beyond historical KV-writing: gradients pass through generated latents into the computations that produced them, linking the generation of earlier context to its use in later predictions. To make this connected training memory-efficient, we develop shortcut gradient replay, which recovers cross-chunk gradients without retaining the full rollout computation graph.
Integrated with distribution matching distillation, Connected Self Forcing trains historical chunks according to both their direct supervision and their contribution to subsequent generation. Experiments on autoregressive video generation show improvements in long-horizon visual quality and temporal consistency, without changing the inference procedure. Additional materials on the \textcolor{magenta}{\href{https://eastbeanzhang.github.io/CSF/}{project page}}.

\end{abstract}

%% file: sec/1_intro.tex
\section{Introduction}
\label{sec:intro}

Recent video generation has advanced rapidly in visual fidelity, motion realism,
and text--video alignment~\citep{Wan,seedance,minimaxh3}, with growing
applications~\citep{VideoPolicy,Emosh,Vace}. Yet many high-quality models still
rely on finite-window bidirectional processing, making them less suitable for
streaming, interactive, and long-horizon generation. Causal autoregressive
generation offers a natural alternative by generating frames or chunks
sequentially from previously generated context, and has therefore been
increasingly explored for long-form video generation and interactive world
modeling~\citep{CausVid,GameNGen,Vid2World,Xworld}.

\begin{figure}[ht]
  \centering
  \includegraphics[width=0.94\textwidth]{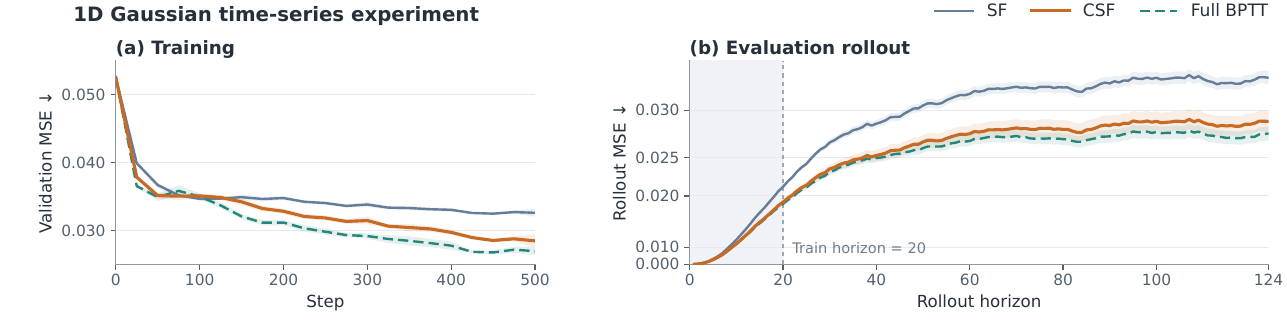}
    \caption{A controlled 1D Gaussian time-series toy experiment. We compare
    Self Forcing (SF), Connected Self Forcing (CSF), and Full BPTT under
    matched autoregressive training conditions. Given the initial context and
    known external inputs at each step, the target trajectory is deterministic,
    allowing MSE evaluation against ground truth. (a) Validation
    MSE. (b) Evaluation rollout MSE.}
  \label{fig:motivation}
\end{figure}

Training causal video generators, however, faces a mismatch between training and
inference: teacher forcing relies on ground-truth history, while Diffusion
Forcing improves robustness by exposing the model to noisy
histories~\citep{DF}. Self Forcing~\citep{SF} goes further by directly training
on the model's own autoregressive rollouts, better aligning training with
inference. To keep such sequential self-rollout tractable, however, previously
generated history is detached when reused by later chunks.

Each generated chunk in an autoregressive rollout serves as both an output and causal context for later predictions. Self Forcing applies a video-level DMD objective to the generated rollout, giving every chunk a direct output gradient~\citep{DMD,DMD2}. However, detaching reused history blocks gradients from later predictions through historical KV states to the computation that generated earlier chunks. The forward dependency remains, but this backward path is absent (Fig.~\ref{fig:teaser}).
To isolate this effect, we use a simplified 1D Gaussian autoregressive
toy experiment (Fig.~\ref{fig:motivation}). Although far simpler than video
generation, it shows SF's long-horizon validation error leveling off while CSF
continues improving toward Full BPTT. CSF also approaches Full BPTT's final
rollout accuracy beyond the training horizon.
These observations motivate the use of future-to-history gradients without
retaining the full rollout graph.

Motivated by these observations, we introduce Connected Self Forcing (CSF),
which restores future-to-history gradient flow without retaining the full
autoregressive computation graph. CSF retains the detached self-rollout of Self Forcing and introduces
Shortcut Gradient Replay. CSF recovers selected cross-chunk derivative paths
rather than the full-sequence gradient. Specifically, a later chunk is replayed with its
historical KV states treated as differentiable, allowing the downstream gradient
to propagate through the KV-writing process and back to the earlier chunk that
produced those states. By replaying only the required computations rather than
backpropagating through the entire rollout, CSF trades recomputation for
substantially lower activation memory. Importantly, CSF adds no model parameters
and leaves inference unchanged.

Experiments show that CSF improves long-horizon visual quality
and temporal consistency. The improvements are reflected in better imaging quality and stronger background and subject consistency, while qualitative results show
reduced subject and scene drift over extended rollouts (Fig.~\ref{fig:teaser}).

Our contributions are as follows:
\textbf{1)} We identify a forward--backward gap in Self Forcing: generated
chunks affect future predictions in the forward rollout, while detachment
blocks gradient paths from later generation computations into earlier context generation.
\textbf{2)} We propose Connected Self Forcing (CSF) with Shortcut Gradient
Replay, reconnecting cross-chunk gradient paths beyond the direct output gradients of DMD, without retaining the full computation graph or changing inference.
\textbf{3)} We demonstrate improved long-horizon generation across 60- and
240-second evaluations, with stronger visual quality and temporal consistency.

%% file: sec/2_related.tex
\section{Related Work}
\label{sec:related}

\noindent\textbf{Bidirectional video generation.}
Video generation has evolved from early diffusion extensions of image models to large-scale foundation models~\citep{VDM,VideoLDM,MovieGen,Wan}. VideoLDM~\citep{VideoLDM} extends pretrained latent diffusion with temporal modeling for video generation. Other early works explore image--video priors, motion modules, and decomposed noise modeling~\citep{VDM,MakeAVideo,AnimateDiff,VideoFusion}, while hierarchical and cascaded designs extend temporal range and spatial resolution~\citep{ImagenVideo,LVDM,LaVie,Show1}. Lumiere~\citep{Lumiere} instead generates the full temporal duration in a single Space-Time U-Net pass. More recent systems explore improved conditioning, transformer architectures, and large-scale data and model scaling~\citep{VideoCrafter1,VideoCrafter2,SVD,Latte,WALT,OpenSora,CogVideoX,HunyuanVideo,MovieGen,Wan}. Such bidirectional computation remains less suitable for low-latency streaming~\citep{CausVid}.

\noindent\textbf{Autoregressive video generation.}
Autoregressive models generate frames or chunks from prior generated
content~\citep{SF,MAGI1,LongLive}. Earlier approaches model tokenized video with
transformers~\citep{VideoGPT,TATS,CogVideo}; VideoPoet~\citep{VideoPoet} extends
decoder-only autoregression to multimodal generation, while
Phenaki~\citep{Phenaki} combines a causal tokenizer with masked-token prediction.
Diffusion Forcing~\citep{DF} generalizes causal sequence modeling to
independently noised continuous tokens. Recent approaches use increasing per-chunk noise~\citep{MAGI1}, Diffusion
Forcing~\citep{SkyReelsV2}, or temporal-pyramid history
compression~\citep{PyramidFlow}. Few-step generation builds on progressive
distillation, consistency models, and distribution matching~\citep{ProgDistill,CM,DMD,DMD2}; CausVid~\citep{CausVid} extends distribution matching to few-step causal video generation.

Train--test mismatch has motivated scheduled sampling and Professor Forcing~\citep{SchedSampling,ProfessorForcing}. Self Forcing~\citep{SF} directly trains on self-generated history with a video-level objective, while gradient truncation keeps training tractable. Subsequent works improve few-step causal initialization~\citep{CausalForcing,CausalForcingPP} and long-horizon training~\citep{RollingForcing,SFPP}.

Concurrent with our work, Self Gradient Forcing~\citep{SGF} uses future losses
to train context KV memory writing while keeping generated context stop-gradient,
whereas Vidu S2's Self-Replay Forcing~\citep{SRF} enables cross-block gradients within a
differentiable replay of re-noised self-generated trajectories. Our method instead
reconnects cross-chunk gradient paths through historical KV writing and
earlier-chunk generation via Shortcut Gradient Replay.

\noindent\textbf{History modeling for long video generation.}
Long-video generation also requires useful history under bounded context.
Training-free extensions use temporal co-denoising, noise rescheduling with windowed
attention, and global/local spectral blending~\citep{GenLVideo,FreeNoise,FreeLong}.
FIFO-Diffusion~\citep{FIFO} performs diagonal queue-based denoising,
RIFLEx~\citep{RIFLEx} adjusts positional-embedding frequency to suppress repetition,
and Ouroboros-Diffusion~\citep{Ouroboros} combines cross-frame attention and recurrent guidance.

Memory mechanisms preserve or retrieve context through previous-chunk attention, frame
sinks with KV recaching, importance-based allocation, and long-context teacher
supervision~\citep{StreamingT2V,LongLive,FramePack,ContextForcing}. More structured
designs include geometry-indexed memory, role-based memory decomposition, distant-history
retrieval, and gated recall~\citep{MemoryForcing,RelaxForcing,RingForcing,TetherCache}.
Ring Forcing~\citep{RingForcing} enforces distant-history retrieval via ring
training and scales memory with compression and sparse RoPE.

Other methods modify temporal supervision and guidance. History-Guided Video Diffusion~\citep{HistoryGuidance} supports flexible history conditioning;
LongVie~\citep{LongVie} improves cross-clip control and consistency, with
LongVie 2~\citep{LongVie2} adding history-context guidance. Video-Mirai~\citep{VideoMirai} distills future-aware features from a
frozen non-causal foresight encoder into causal states, whereas Next
Forcing~\citep{NextForcing} adds predictors over multiple future horizons. CSF instead propagates gradients from later-chunk losses into earlier context generation.

%% file: sec/3_method.tex
\section{Method}
\label{sec:method}

\begin{figure*}[t]
  \centering
  \includegraphics[width=\textwidth]{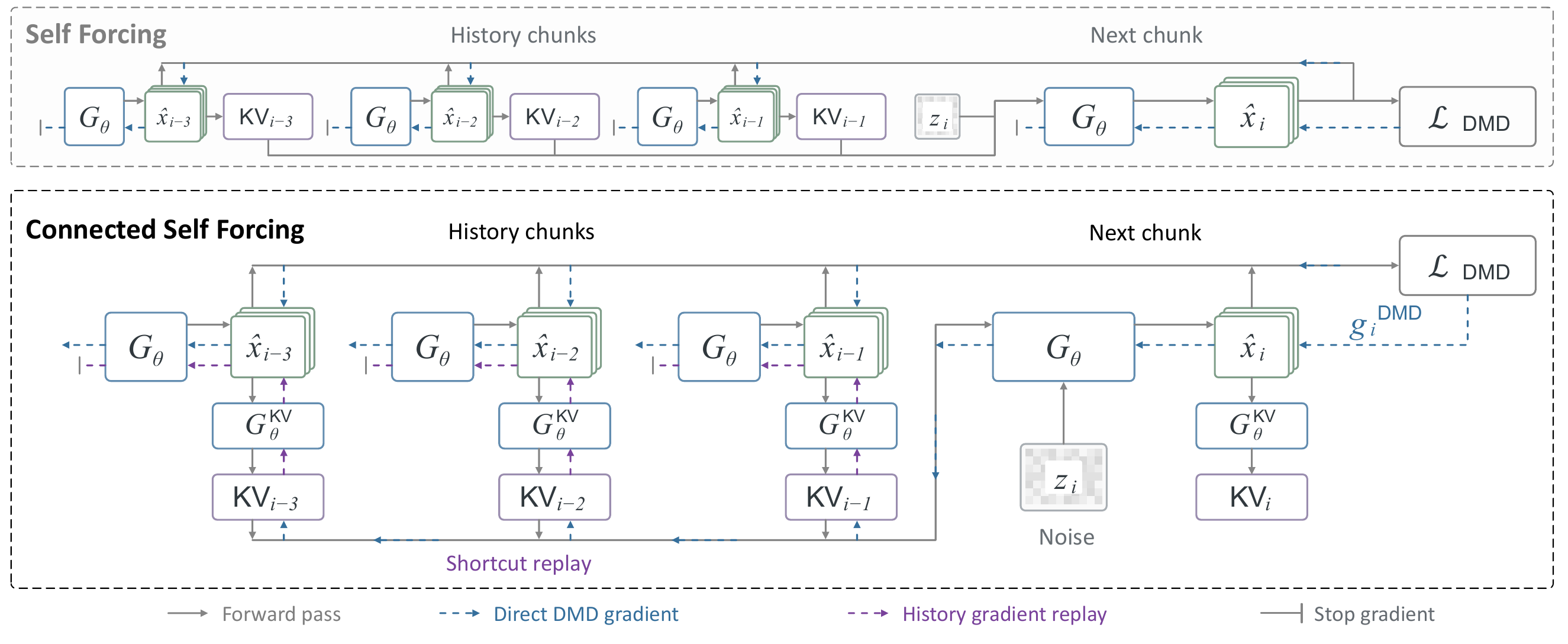}
  \caption{\textbf{Overview of Connected Self Forcing.} Self Forcing uses
    detached autoregressive rollouts, where each generated chunk $\hat{x}_i$
    receives only direct DMD supervision. Connected Self Forcing restores
    cross-chunk feedback through \emph{Shortcut Gradient Replay}: gradients from
    a later prediction are replayed to historical KV states $\mathrm{KV}_j$,
    then propagated through the KV writer $G_\theta^{\mathrm{KV}}$ and generator
    $G_\theta$, which share parameters $\theta$. Replay stops at earlier
    generated history to avoid recursive backpropagation through the earlier generation branches. This trains each chunk both for its own generation quality and as
    context for subsequent generation, without adding parameters or changing
    inference.}
  \label{fig:pipeline}
\end{figure*}

In this section, we present Connected Self Forcing (CSF), illustrated in Fig.~\ref{fig:pipeline}. We first
review Self Forcing in Sec.~\ref{sec:method:background} and formulate the
cross-chunk gradient assignment problem under detached autoregressive training in
Sec.~\ref{sec:method:assign}. We then introduce Shortcut Gradient Replay
(Sec.~\ref{sec:method:replay}), our mechanism for recovering this missing
training signal, and finally describe the complete CSF training procedure
(Sec.~\ref{sec:method:training}), summarized in Algorithm~\ref{alg:csf}.

\subsection{Autoregressive Video Generation with Self Forcing}
\label{sec:method:background}

In autoregressive video generation, each chunk is generated conditioned on
previously generated history. Self Forcing~\citep{SF} matches this
inference-time conditioning during training by rolling out on the model's own
generated history. Let $G_\theta$ generate chunk $i$, and let
$G_\theta^{\mathrm{KV}}$ denote the KV-writing computation of the same
generator:
\begin{equation}
\begin{aligned}
  \hat{\mathbf{x}}_i
  &= G_\theta(\mathbf{z}_i;
  \operatorname{sg}(\mathcal{C}_{<i}), \mathbf{c}),\\
  \mathcal{C}_i
  &= G_\theta^{\mathrm{KV}}(\hat{\mathbf{x}}_i;
  \operatorname{sg}(\mathcal{C}_{<i}), \mathbf{c}).
\end{aligned}
\label{eq:sf_rollout}
\end{equation}
Here, $\mathbf{z}_i$ is the noisy latent, $\mathbf{c}$ the text condition, and
$\mathcal{C}_{<i}$ the KV cache from preceding generated chunks. The second
line re-encodes $\hat{\mathbf{x}}_i$ at the clean context timestep to update
the causal cache, yielding the causal rollout
$\hat{\mathbf{x}}_1 \rightarrow \mathcal{C}_1 \rightarrow \hat{\mathbf{x}}_2
\rightarrow \mathcal{C}_2 \rightarrow \cdots$. To keep the sequential rollout
tractable, reused history and its KV cache are detached, so only each chunk's
local computation graph is retained.

Following prior autoregressive distillation methods~\citep{CausVid,SF}, the
self-generated rollout $\hat{\mathbf{X}}=[\hat{\mathbf{x}}_1,\ldots,\hat{\mathbf{x}}_N]$
is optimized with Distribution Matching Distillation (DMD)~\citep{DMD,DMD2}.
DMD evaluates re-noised generated samples using a pretrained real-score model
and an online fake-score model that tracks the generator distribution,
yielding the direct output gradient
\begin{equation}
  \mathbf{g}^{\mathrm{DMD}}
  =
  \frac{\partial \mathcal{L}_{\mathrm{DMD}}}
  {\partial \hat{\mathbf{X}}}.
\label{eq:dmd-grad}
\end{equation}
Thus, every generated chunk receives direct DMD supervision, while the
limitation we study arises from the detached dependencies between chunks.

\subsection{Cross-Chunk Gradient Assignment}
\label{sec:method:assign}

In autoregressive generation, each historical chunk plays two roles: it is both a
generated output and part of the causal context for subsequent predictions.
Distribution matching directly supervises the former by providing a gradient to
every generated chunk. However, once a generated chunk is reused as detached
history, its downstream influence on later predictions no longer contributes to
its own supervision. This creates an asymmetry: the dependency is present during
the forward rollout, but absent from the backward learning signal.

Consider a historical chunk $\hat{\mathbf{x}}_j$ and a later chunk
$\hat{\mathbf{x}}_i$, where $j < i$. When a causal path exists, their
computational dependency is
\begin{equation}
  \hat{\mathbf{x}}_j \xrightarrow{\mathcal{P}_{j\rightarrow i}} \hat{\mathbf{x}}_i,
  \label{eq:causal-path}
\end{equation}
where $\mathcal{P}_{j\rightarrow i}$ collectively represents all differentiable
causal paths from $\hat{\mathbf{x}}_j$ to $\hat{\mathbf{x}}_i$. These paths may
involve intermediate generated chunks, direct access to historical
representations, or successive updates of a persistent memory state. This
formulation does not assume a particular history representation or propagation
mechanism.

Although $\hat{\mathbf{x}}_j$ already receives its direct DMD gradient, gradient
truncation removes the additional supervision induced by its effect on later
predictions. We refer to this missing backward dependency as the
\emph{cross-chunk gradient gap}. With model parameters and rollout randomness
held fixed, the total Jacobian over these paths before cross-chunk gradient
truncation and the resulting cross-chunk gradient are
\begin{equation}
\begin{aligned}
  \mathbf{J}_{\mathcal{P}_{j\rightarrow i}}
  &\triangleq \frac{\mathrm{d}\hat{\mathbf{x}}_i}
                    {\mathrm{d}\hat{\mathbf{x}}_j},\\
  \mathbf{g}^{\mathrm{cross}}_{i\rightarrow j}
  &= \mathbf{J}_{\mathcal{P}_{j\rightarrow i}}^{\top}
     \mathbf{g}^{\mathrm{DMD}}_i.
\end{aligned}
\label{eq:cross-grad}
\end{equation}
This term assigns later supervision to an earlier generated chunk according to
its downstream influence. It is a retrospective training signal defined on the
same self-generated rollout: later predictions only determine how gradient is
assigned to earlier generated context. Consequently, historical chunks should be
optimized both for their own generation quality and for their role as causal
context for subsequent predictions.

\subsection{Shortcut Gradient Replay}
\label{sec:method:replay}

A straightforward way to obtain the cross-chunk gradient in
Sec.~\ref{sec:method:assign} is to keep the autoregressive computation graph
connected across the entire self-rollout and directly backpropagate through it.
However, each chunk involves multiple denoising and context-writing forwards of a
large video generator. Retaining all intermediate activations across a long
rollout therefore incurs prohibitive memory cost. We instead trade activation
memory for recomputation through gradient replay. During the forward rollout, we
discard the serial computation graph and retain only the states required to
reproduce each chunk.

A natural replay strategy is to propagate the future gradient backward one chunk
at a time along the autoregressive dependencies: each chunk is replayed to obtain
a gradient for its preceding history, which is then recursively relayed toward
earlier chunks. While this avoids retaining the full graph, the number of
replayed transitions grows with the gradient distance. Moreover, long-range
gradient is repeatedly transformed by the Jacobians of intervening chunks, making
the resulting signal increasingly dependent on the intermediate dynamics.

Our key observation is that the causal attention structure already provides a
direct dependency to retained historical KV states: a later chunk directly reads
the cached keys and values written by preceding chunks. We therefore recover selected
path contributions to Eq.~(\ref{eq:cross-grad}) through retained historical KV
states, bypassing intervening generation transitions without backpropagating
through the full rollout graph. Specifically, we replay chunk
$i$ while treating the selected historical cache $\mathcal{C}_j$ as a
differentiable leaf and keeping the remaining history detached. Given the direct
DMD gradient $\mathbf{g}_i^{\mathrm{DMD}}$, this replay yields
\begin{equation}
  \mathbf{g}_{\mathcal{C}_j}
  = \left( \frac{\partial \hat{\mathbf{x}}_i}{\partial \mathcal{C}_j} \right)^{\!\top}
    \mathbf{g}_i^{\mathrm{DMD}}.
  \label{eq:replay-kv}
\end{equation}
This establishes a direct shortcut from the later prediction to the historical KV
state without propagating the gradient through the intervening chunks. The
resulting gradient on $\mathcal{C}_j$ is then backpropagated through the
context-writing computation of chunk $j$. This replay both updates the
context-writing parameters and produces a gradient on the recorded historical
output,
\begin{equation}
  \mathbf{g}_{\hat{\mathbf{x}}_j}^{\mathrm{cross}}
  = \left( \frac{\partial \mathcal{C}_j}{\partial \hat{\mathbf{x}}_j} \right)^{\!\top}
    \mathbf{g}_{\mathcal{C}_j}.
  \label{eq:replay-output}
\end{equation}
We then replay the primary generation of chunk $j$ using
$\mathbf{g}_{\hat{\mathbf{x}}_j}^{\mathrm{cross}}$ as its output gradient,
allowing the cross-chunk feedback to update the computation that produced the
historical chunk. This generation feedback terminates at the targeted chunk:
earlier generated outputs are not recursively replayed through their generation
branches. In parallel, the KV-writing replay can propagate its incoming KV
gradient to earlier KV writers while keeping each writer's generated-latent
input detached, preventing this writer-only feedback from re-entering earlier
generation branches. We call this procedure
\emph{Shortcut Gradient Replay}, as it directly connects later predictions to
selected historical chunks without recursively replaying the intervening
generation trajectory.

\subsection{Training}
\label{sec:method:training}

\newcounter{algcsf}
\begin{wrapfigure}[20]{R}{0.5\textwidth}
\vspace{-0.6\baselineskip}
\hrule height 0.8pt
\vspace{3pt}
{\scriptsize
\refstepcounter{algcsf}\label{alg:csf}%
\noindent\textbf{Algorithm~\thealgcsf:} Connected Self Forcing Training
Procedure\par
\vspace{2pt}
\hrule height 0.4pt
\vspace{2pt}
\begin{algorithmic}[1]
\Require $G_\theta$, $G_\theta^{\mathrm{KV}}$, $S_{\mathrm{real}}$,
$S_{\mathrm{fake},\phi}$, $\lambda_{\mathrm{cross}}$

\For{$i=1,\ldots,N$} \Comment{detached self-rollout}
    \State Generate $\hat{\mathbf{x}}_i$ via standard denoising with
    $\operatorname{sg}(\mathcal{C}_{<i})$
    \State Write
    $\mathcal{C}_i \gets
    G_\theta^{\mathrm{KV}}(\hat{\mathbf{x}}_i;
    \operatorname{sg}(\mathcal{C}_{<i}),\mathbf{c})$
    and save replay states
\EndFor

\State Compute
$\mathbf{g}^{\mathrm{DMD}}
=\partial\mathcal{L}_{\mathrm{DMD}}/\partial\hat{\mathbf{X}}$
and obtain $\{\mathbf{g}_i^{\mathrm{DMD}}\}_{i=1}^{N}$
\State Initialize $\mathbf{g}_{\mathcal{C}_j}\gets 0$

\For{$i=N,\ldots,1$} \Comment{reverse replay}
    \State Replay chunk $i$ with historical KV states differentiable
    \State Backpropagate $\mathbf{g}_i^{\mathrm{DMD}}$ and collect
    \State \hspace{\algorithmicindent}
    $\displaystyle
    \mathbf{g}_{\mathcal{C}_j} \mathrel{+}=
    \lambda_{\mathrm{cross}}\left(
    \frac{\partial\hat{\mathbf{x}}_i}
         {\partial\mathcal{C}_j}
    \right)^{\!\top}
    \mathbf{g}_i^{\mathrm{DMD}},
    \quad i\in\mathcal{F}(j)$

    \If{$\mathbf{g}_{\mathcal{C}_i}\neq 0$}
        \State Replay $G_\theta^{\mathrm{KV}}$ and earlier KV writers
        \Statex \hspace{24pt}with earlier generated outputs detached
        \State \hspace{\algorithmicindent}
        $\displaystyle
        \mathbf{g}_{\hat{\mathbf{x}}_i}^{\mathrm{cross}}
        =
        \left(
        \frac{\partial\mathcal{C}_i}
             {\partial\hat{\mathbf{x}}_i}
        \right)^{\!\top}
        \mathbf{g}_{\mathcal{C}_i}$
        \State Replay chunk $i$ with detached history using
        $\mathbf{g}_{\hat{\mathbf{x}}_i}^{\mathrm{cross}}$
    \EndIf
\EndFor

\State Update $\theta$ with accumulated gradients;
update $\phi$ with the standard fake-score objective
\end{algorithmic}
\vspace{2pt}
\hrule height 0.4pt
}
\vspace{-0.2\baselineskip}
\end{wrapfigure}

During training, the gradient assigned to each historical chunk $j$ combines its
direct DMD gradient with cross-chunk feedback from later chunks influenced by its
KV state:
\begin{equation}
  \tilde{\mathbf{g}}_j = \mathbf{g}_j^{\mathrm{DMD}}
  + \lambda_{\mathrm{cross}} \sum_{i \in \mathcal{F}(j)}
    \widehat{\mathbf{g}}_{i \rightarrow j}^{\mathrm{cross}},
  \label{eq:training-grad}
\end{equation}
where $\widehat{\mathbf{g}}_{i\rightarrow j}^{\mathrm{cross}}$ denotes the
contribution of chunk $i$ to the replayed output gradient in
Eq.~(\ref{eq:replay-output}), $\mathcal{F}(j)$ contains
later chunks that directly read $\mathcal{C}_j$ within the considered history
range, and
$\lambda_{\mathrm{cross}}=1$ by default scales added feedback to both KV-writing
and historical-generation parameter gradients.

Algorithm~\ref{alg:csf} summarizes the resulting procedure.
For each generator update, we first perform a detached autoregressive
self-rollout and record the states required for replay. The generated video is
then evaluated with the standard DMD objective to obtain chunk-wise output
gradients. Shortcut Gradient Replay reconstructs the corresponding cross-chunk
feedback, which is accumulated together with the direct DMD gradients on the
shared generator parameters before a single optimizer update.
We use DMD here; CSF's replay can also use output gradients from other
differentiable losses.

Connected Self Forcing modifies only the generator-side gradient propagation
during training. The real-score model remains fixed, while the fake-score model is
updated with the standard DMD denoising objective and does not require gradient
replay. The method introduces no additional model parameters. At inference time,
all replay operations are removed, and the generator follows the same causal
autoregressive rollout and KV-cache mechanism as the original Self Forcing model.

%% file: sec/4_experiment.tex
\section{Experiments}
\label{sec:experiments}

\subsection{Experimental Setup}
\label{sec:exp_setup}

\noindent\textbf{Training Details.}
We perform text-to-video DMD post-training using the released filtered and
extended VidProM prompt set~\citep{VidProM}, without loading additional real
videos during this stage. Our main experiments use Wan2.1-T2V-1.3B~\citep{Wan}
as the causal generator. Following Self Forcing~\citep{SF}, we optimize the
generator with DMD~\citep{DMD,DMD2}, using a frozen Wan2.1-T2V-14B real-score
model and a trainable Wan2.1-T2V-1.3B fake-score model. We use four denoising
steps and primarily study chunk-wise generation, where each 21-latent-frame
training sequence is divided into seven chunks. Training uses AdamW with a
global batch size of 8 for 1,200 iterations.

\noindent\textbf{Baselines.}
We primarily compare Connected Self Forcing (CSF) with Self Forcing
(SF)~\citep{SF} and the concurrent Self Gradient Forcing
(SGF)~\citep{SGF}. For paired comparison, we train both baselines following
their released implementations and training recipes. Within each controlled
comparison, methods use matched
backbones, causal initialization, training prompts, training budgets, and
inference configurations. This allows us to focus the comparison on how
self-generated history is optimized during autoregressive training.

\noindent\textbf{Evaluation Protocol.}
For long-horizon evaluation, we generate approximately 60-second videos from
40 VBench-Long prompts~\citep{VBenchPP} and approximately 240-second videos
from a fixed 128-prompt subset of Movie Gen Video Bench~\citep{MovieGen}.
We evaluate the EMA generator checkpoints with matched four-step causal
sampling, inference seeds, resolution, and KV-cache configurations. Following
VBench-Long~\citep{VBenchPP}, we report seven quality and temporal dimensions:
aesthetic quality, background consistency, dynamic degree, imaging quality,
motion smoothness, subject consistency, and temporal flickering. Further
implementation and evaluation details are provided in the supplementary
material.

\subsection{Quantitative Results}
\label{sec:quantitative}

Tables~\ref{tab:chunkwise_60s} and~\ref{tab:chunkwise_240s} report
long-horizon chunk-wise generation results at 60 and 240 seconds.
We report \textbf{Avg.(6)}, our mean of six quality and consistency metrics,
excluding Dynamic Degree.

\begin{table*}[t]
\centering
\caption{60-second chunk-wise generation on VBench-Long.
Avg.(6) averages Aesthetic, Background, Imaging, Motion Smoothness,
Subject Consistency, and Flickering; Dynamic Degree is reported separately.
Best and second-best values are highlighted except for Dynamic Degree.}
\label{tab:chunkwise_60s}
\small
\setlength{\tabcolsep}{3.0pt}
\renewcommand{\arraystretch}{0.95}

\begin{tabular}{lccc ccc ccc}
\toprule
& \multicolumn{3}{c}{Causal CD Init}
& \multicolumn{3}{c}{TF Init}
& \multicolumn{3}{c}{Causal ODE Init} \\
\cmidrule(lr){2-4}
\cmidrule(lr){5-7}
\cmidrule(lr){8-10}

Metric
& SF & SGF & \textbf{CSF (Ours)}
& SF & SGF & \textbf{CSF (Ours)}
& SF & SGF & \textbf{CSF (Ours)} \\
\midrule

Aesthetic
& 0.5473 & \underline{0.5592} & \textbf{0.5659} & \underline{0.5990} & 0.5917 & \textbf{0.6152} & 0.5264 & \underline{0.5772} & \textbf{0.5804} \\

Background
& 0.9567 & \underline{0.9583} & \textbf{0.9597} & \underline{0.9601} & 0.9543 & \textbf{0.9721} & 0.9533 & \underline{0.9625} & \textbf{0.9647} \\

Imaging
& 0.6939 & \underline{0.7031} & \textbf{0.7085} & 0.7046 & \underline{0.7152} & \textbf{0.7286} & 0.7123 & \textbf{0.7224} & \underline{0.7191} \\

Motion
& \textbf{0.9884} & \underline{0.9849} & 0.9848 & \underline{0.9846} & 0.9764 & \textbf{0.9906} & 0.9817 & \underline{0.9874} & \textbf{0.9896} \\

Subject
& 0.9656 & \underline{0.9665} & \textbf{0.9707} & \underline{0.9695} & 0.9671 & \textbf{0.9821} & 0.9563 & \textbf{0.9733} & \underline{0.9727} \\

Flickering
& \textbf{0.9704} & \underline{0.9657} & 0.9637 & \underline{0.9702} & 0.9442 & \textbf{0.9791} & 0.9611 & \underline{0.9729} & \textbf{0.9733} \\

Dynamic
& 0.4435 & 0.6113 & 0.5976 & 0.6992 & 0.8653 & 0.4960 & 0.5742 & 0.5565 & 0.4427 \\

\midrule

Avg.(6)
& 0.8537 & \underline{0.8563} & \textbf{0.8589} & \underline{0.8647} & 0.8582 & \textbf{0.8780} & 0.8485 & \underline{0.8660} & \textbf{0.8666} \\
\bottomrule
\end{tabular}
  \vspace{-5pt}
\end{table*}

\begin{table*}[t]
\centering
\caption{240-second chunk-wise generation on the fixed 128-prompt MovieGen Video Bench subset.
Metrics and formatting follow Table~\ref{tab:chunkwise_60s}.}
\label{tab:chunkwise_240s}
\small
\setlength{\tabcolsep}{3.0pt}
\renewcommand{\arraystretch}{0.95}

\begin{tabular}{lccc ccc ccc}
\toprule
& \multicolumn{3}{c}{Causal CD Init}
& \multicolumn{3}{c}{TF Init}
& \multicolumn{3}{c}{Causal ODE Init} \\
\cmidrule(lr){2-4}
\cmidrule(lr){5-7}
\cmidrule(lr){8-10}

Metric
& SF & SGF & \textbf{CSF (Ours)}
& SF & SGF & \textbf{CSF (Ours)}
& SF & SGF & \textbf{CSF (Ours)} \\
\midrule

Aesthetic
& 0.5138 & \underline{0.5220} & \textbf{0.5449} & \underline{0.5639} & 0.5626 & \textbf{0.5948} & 0.5059 & \textbf{0.5357} & \underline{0.5260} \\

Background
& 0.9524 & \underline{0.9539} & \textbf{0.9563} & \underline{0.9554} & 0.9512 & \textbf{0.9673} & 0.9536 & \underline{0.9559} & \textbf{0.9607} \\

Imaging
& \underline{0.6700} & 0.6607 & \textbf{0.6938} & 0.6937 & \underline{0.7030} & \textbf{0.7161} & 0.6907 & \underline{0.6965} & \textbf{0.7004} \\

Motion
& \textbf{0.9887} & 0.9845 & \underline{0.9850} & \underline{0.9835} & 0.9761 & \textbf{0.9893} & 0.9838 & \underline{0.9847} & \textbf{0.9895} \\

Subject
& \underline{0.9582} & 0.9581 & \textbf{0.9666} & \underline{0.9647} & 0.9619 & \textbf{0.9786} & 0.9598 & \underline{0.9658} & \textbf{0.9688} \\

Flickering
& \textbf{0.9691} & 0.9645 & \underline{0.9656} & \underline{0.9659} & 0.9421 & \textbf{0.9747} & 0.9651 & \underline{0.9690} & \textbf{0.9734} \\

Dynamic
& 0.4126 & 0.7056 & 0.6899 & 0.7131 & 0.9200 & 0.5539 & 0.5506 & 0.6306 & 0.5003 \\

\midrule

Avg.(6)
& \underline{0.8420} & 0.8406 & \textbf{0.8520} & \underline{0.8545} & 0.8495 & \textbf{0.8701} & 0.8431 & \underline{0.8513} & \textbf{0.8531} \\
\bottomrule
\end{tabular}
  \vspace{-5pt}
\end{table*}

\begin{figure*}[!t]
  \centering
  \includegraphics[width=\textwidth]{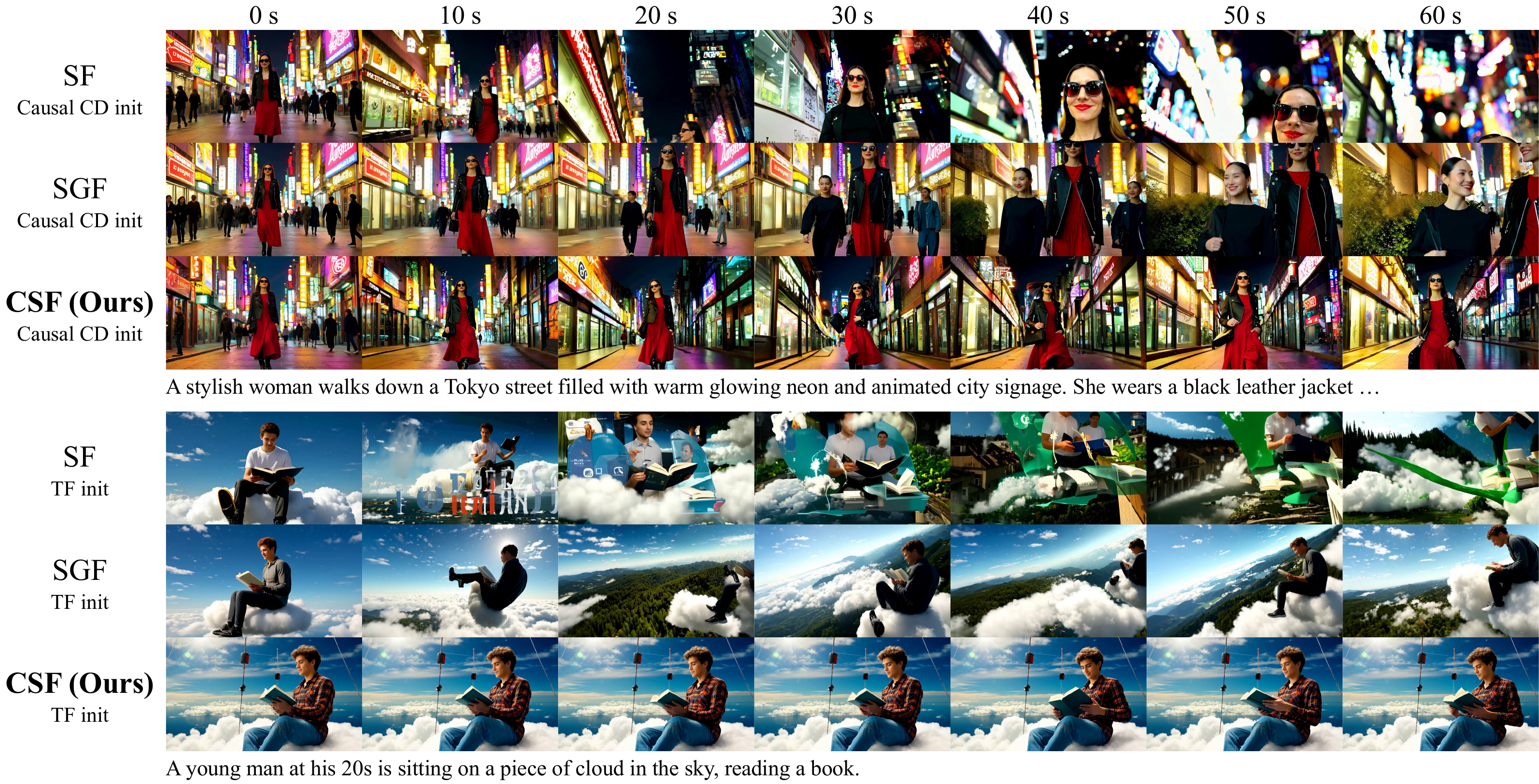}
  \caption{Qualitative comparison of 60-second chunk-wise autoregressive
    generation: SF, SGF, and CSF on VBench-Long prompts under
    Causal CD and TF initialization. CSF more consistently preserves subject
    appearance and scene composition throughout the rollout across both
    settings.}
  \label{fig:qualitative_60s}
  \vspace{-5pt}
\end{figure*}

\begin{figure*}[!t]
  \centering
  \includegraphics[width=\textwidth]{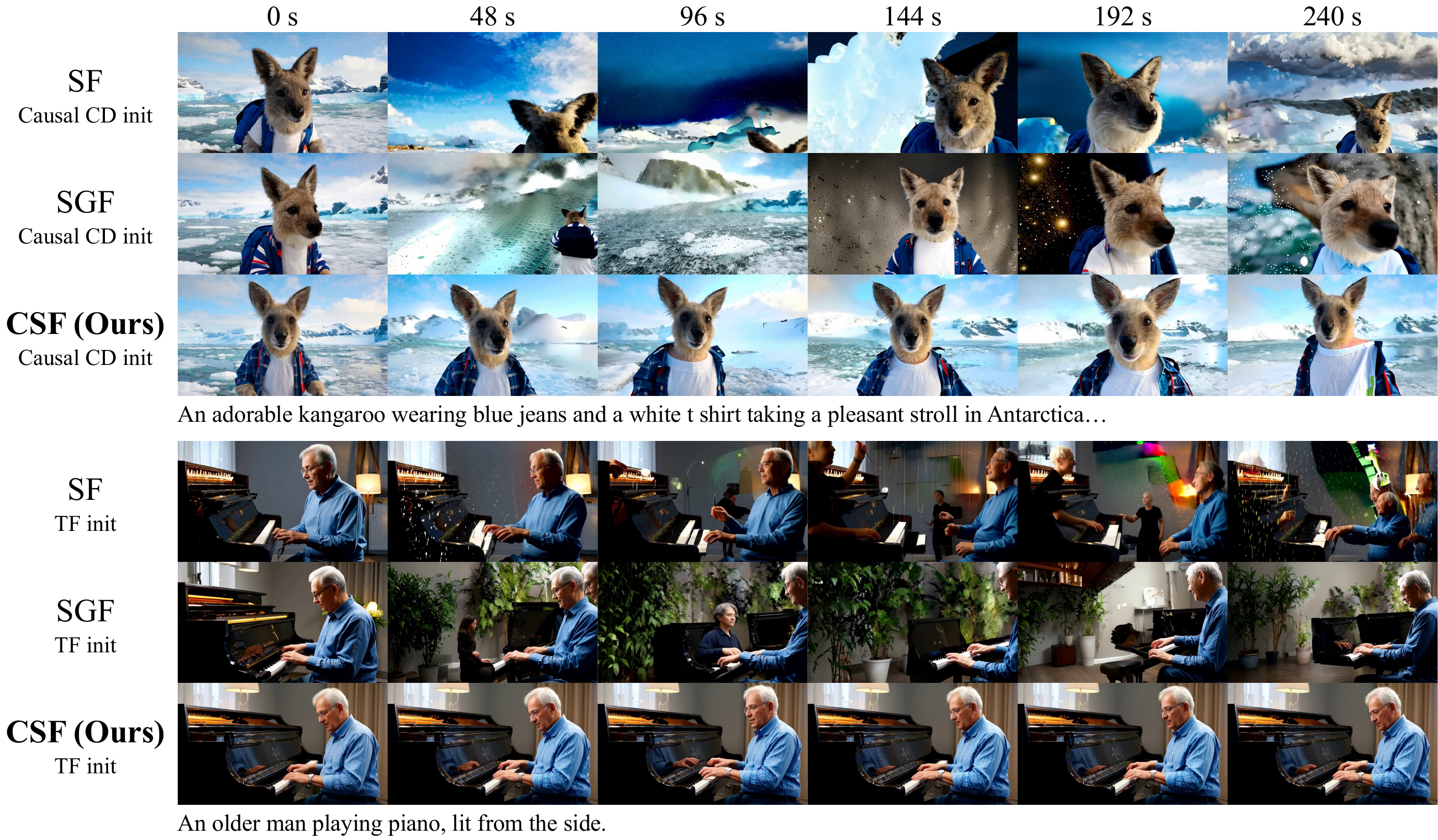}
  \caption{Qualitative comparison of 240-second chunk-wise autoregressive
    generation. We compare SF, SGF, and CSF on MovieGen-128 prompts under
    Causal CD and TF initialization. Over four-minute rollouts, CSF better
    preserves subject identity and scene semantics, with less accumulated
    drift in both the kangaroo and pianist examples.}
  \label{fig:qualitative_240s}
  \vspace{-5pt}
\end{figure*}
CSF achieves the highest Avg.(6) point estimate in all six chunk-wise
settings. At 240 seconds, paired bootstrap intervals lie above zero
against SF under all three initializations, and against SGF under
Causal CD and TF (Appendix~\ref{app:paired_bootstrap}). Because prompt sets differ, we compare methods within each duration rather than scores across durations.

The component metrics further show where the improvements arise.
At 60 seconds, CSF achieves the best Aesthetic Quality and Background
Consistency under all three initializations. At 240 seconds, CSF ranks
first in Background Consistency, Imaging Quality, and Subject Consistency
under all three initializations. These improvements are particularly
relevant to long-horizon autoregressive generation, where subject and
scene drift can accumulate throughout the rollout, and are consistent
with the goal of propagating downstream supervision back to historical
generation.

CSF's balance of quality, consistency, and dynamics depends on
initialization: under Causal CD, it improves Avg.(6) with Dynamic Degree
near SGF and above SF at both horizons; under TF and Causal ODE, dynamics
are below both baselines. Thus, we interpret Avg.(6) alongside Dynamic
Degree without claiming uniform improvement.
Efficiency and memory are reported in Appendix~\ref{app:training_efficiency}.

\subsection{Qualitative Comparison}
\label{sec:qualitative}
We further compare long-horizon visual consistency by tracking how subjects
and scenes evolve throughout autoregressive rollouts.
Figures~\ref{fig:qualitative_60s} and~\ref{fig:qualitative_240s} show
uniformly spaced snapshots from 60-second and 240-second generations,
respectively.

For the 60-second examples, CSF shows relatively stable subject appearance and
scene composition under both Causal CD and TF initialization. In the
Tokyo-street example, SF gradually shifts toward closer views of the subject,
while SGF shows changes in both the subject and background. CSF retains a
similar subject appearance and neon-street setting across the sampled
timestamps. In the cloud-reading example, SF and SGF show larger changes in
the surrounding scene over time, whereas the CSF sample remains closer to the
initial subject and cloud setting.

Likewise, in the 240-second Antarctica example, SF and SGF change subject
scale, appearance, and background; CSF stays closer to the initial kangaroo
and snowy setting. For the pianist, CSF keeps subject, piano, and indoor
composition similar across sampled timestamps; SF and SGF show larger scene
and appearance changes. In both long rollouts, CSF shows less visible subject
and scene drift, consistent with the motivation to reconnect downstream
supervision to historical generation.

\subsection{Ablation Study}
\label{sec:ablation}

We ablate cross-chunk gradient propagation in the 60-second chunk-wise Causal
CD setting. \textbf{Serial recursion} recursively propagates accumulated future gradients backward, one adjacent chunk at a time. \textbf{Full BPTT}
replays the selected denoising exit with a connected graph across the autoregressive history. \textbf{CSF (Ours)} instead
uses shortcut replay and
terminates feedback at the targeted historical chunk. We also evaluate
\textbf{CSF ($\lambda=0.5$)}, halving the added cross-chunk feedback.

\begin{table}[t]
\centering
\caption{Ablation study on 60-second chunk-wise generation with Causal CD initialization.
Avg.(6) and highlighting conventions follow Table~\ref{tab:chunkwise_60s}.}
\label{tab:ablation}
\small
\setlength{\tabcolsep}{3.0pt}
\renewcommand{\arraystretch}{0.95}

\begin{tabular}{lcccccccc}
\toprule
Method
& Aes.$\uparrow$
& Back.$\uparrow$
& Imag.$\uparrow$
& Mot.$\uparrow$
& Subj.$\uparrow$
& Flick.$\uparrow$
& Dyn.
& Avg.(6)$\uparrow$ \\
\midrule

Full BPTT
& \underline{0.5597} & 0.9562 & 0.6808 & \underline{0.9857} & 0.9661 & \underline{0.9688} & 0.5742 & 0.8529 \\

Serial recursion
& 0.5432 & \textbf{0.9628} & \underline{0.6907} & \textbf{0.9887} & \textbf{0.9753} & \textbf{0.9726} & 0.4032 & \underline{0.8556} \\

CSF ($\lambda=0.5$)
& 0.5471 & 0.9498 & 0.6852 & 0.9769 & 0.9486 & 0.9497 & 0.8484 & 0.8429 \\

\textbf{CSF (Ours)}
& \textbf{0.5659} & \underline{0.9597} & \textbf{0.7085} & 0.9848 & \underline{0.9707} & 0.9637 & 0.5976 & \textbf{0.8589} \\
\bottomrule
\end{tabular}
\vspace{-5pt}
\end{table}

\begin{figure*}[t]
  \centering
  \includegraphics[width=\textwidth]{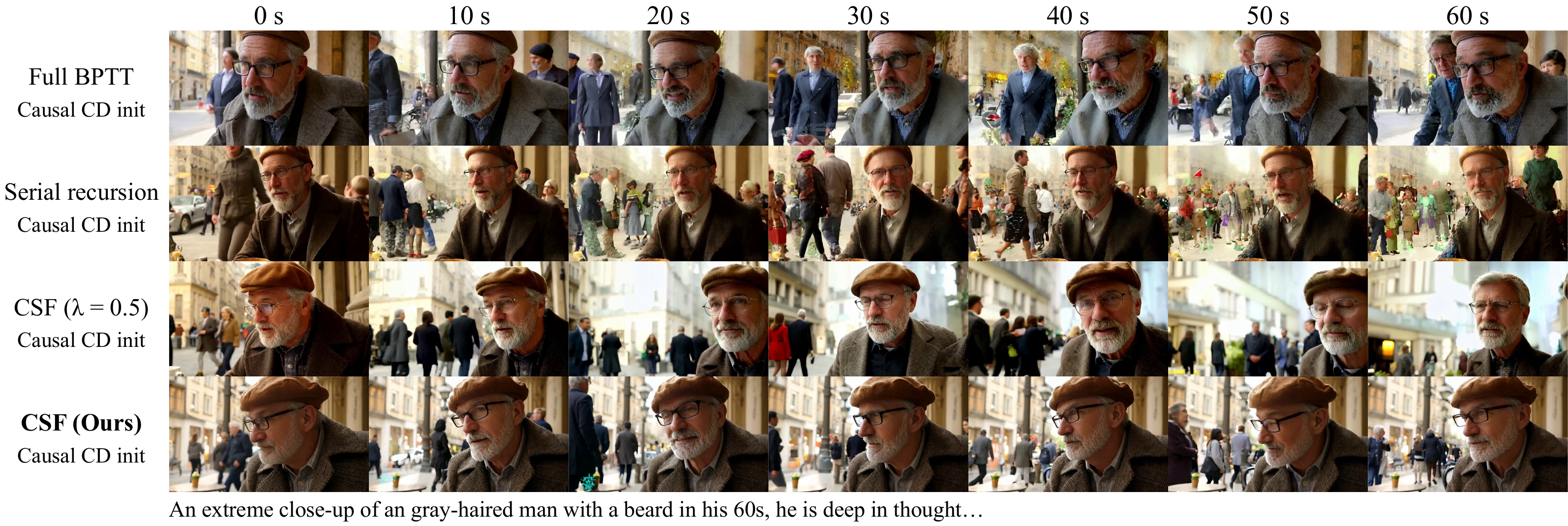}
  \caption{Qualitative ablation on 60-second chunk-wise generation with Causal
    CD initialization.}
  \label{fig:ablation}
  \vspace{-5pt}
\end{figure*}

As shown in Table~\ref{tab:ablation}, Serial recursion remains close to
CSF in quality and consistency metrics, but yields substantially lower
Dynamic Degree. Reducing $\lambda$ to 0.5 substantially increases
Dynamic Degree, while degrading several quality and consistency metrics.
Here, Full BPTT does not outperform CSF on Avg.(6) or Dynamic Degree.
CSF shows a favorable balance of long-horizon quality, consistency, and
motion in this setting.
The qualitative comparison in Fig.~\ref{fig:ablation} shows a similar
pattern over the rollout.

%% file: sec/5_conclusion.tex
\section{Conclusion}
\label{sec:conclusion}

We introduced Connected Self Forcing (CSF) to address missing
future-to-history gradient flow in autoregressive self-rollout training. Its
Shortcut Gradient Replay reconnects downstream supervision to historical
generation without retaining the full autoregressive computation graph or
changing inference. Across several initializations, CSF consistently improves
long-horizon visual quality and temporal consistency in 60- and 240-second
autoregressive videos. By extending gradient flow beyond detached chunk boundaries,
CSF highlights the potential of future-aware training for autoregressive video
generation.

%% file: sec/7_appendix.tex
\clearpage
\begin{center}
    {\Large\bfseries Supplementary Material for Connected Self Forcing}
\end{center}

\appendix

\providecommand{\ToyFigureReference}{Fig.~\ref{fig:motivation}}

\section*{Overview}

\noindent
This supplementary material provides additional details, experiments,
qualitative results, and discussion that complement the main paper.
It is organized as follows:
\begin{itemize}
    \item A \href{https://eastbeanzhang.github.io/CSF/}{supplementary video demo},
    with long-horizon chunk-wise and frame-wise comparisons, in
    Appendix~\ref{app:video_demo}.

    \item Additional implementation and training details in
    Appendix~\ref{app:implementation}.

    \item A controlled Gaussian time-series toy experiment providing
    additional details and analysis for the motivating study in
    Appendix~\ref{sec:toy-concise}.

    \item Further experiments, including training efficiency and memory,
    short-horizon chunk-wise evaluation, additional qualitative results,
    frame-wise experiments, paired bootstrap uncertainty analysis, and gradient-conflict analysis, in
    Appendix~\ref{app:further_experiments}.

    \item Limitations and further discussion in
    Appendix~\ref{app:limitations}.
\end{itemize}

\section{Video Demo}
\label{app:video_demo}

We encourage readers to watch the \href{https://eastbeanzhang.github.io/CSF/}{supplementary video demo}, which complements
the sampled frames in the paper with synchronized excerpts from 60- and
240-second autoregressive generations. The demo compares Connected Self
Forcing (CSF) with Self Forcing (SF)~\citep{SF} and Self Gradient Forcing (SGF)~\citep{SGF} under
matched prompts and inference settings, covering both chunk-wise and frame-wise
generation. Side-by-side clips illustrate how subject appearance, scene
composition, and motion evolve over time, while source-time indicators
explicitly mark the temporal locations of the selected excerpts. The video also
includes a brief four-way qualitative ablation of cross-chunk gradient
propagation under Causal CD initialization. These moving comparisons provide
temporal information that is difficult to capture from isolated frames alone.

\section{Additional Implementation Details}
\label{app:implementation}

\subsection{Training Configuration}
\label{app:training_config}

We use the same filtered and extended VidProM-derived prompt set used by
SGF~\citep{SGF}, containing approximately 250K prompts, for all post-training
experiments. For chunk-wise training, the 21 latent frames are divided into
seven three-frame chunks. We use a bounded causal context consisting of a
three-latent-frame persistent sink and the six most recent latent frames,
corresponding to one sink chunk and two recent chunks.

Following SGF~\citep{SGF}, we adopt a stochastic exit strategy: one of the four
denoising steps is sampled independently for each training video and shared
across all autoregressive chunks in the rollout. The prediction at the sampled
exit, rather than the final clean prediction, is used to construct the causal
history.

We use AdamW with generator and fake-score learning rates of
$2\times10^{-6}$ and $4\times10^{-7}$, respectively,
$\beta=(0,0.999)$, and weight decay $0.01$. The fake-score model is updated
every iteration, while the generator is updated once every five iterations.
We maintain an EMA of the generator with decay $0.99$ and use the EMA weights
for evaluation. Chunk-wise experiments are trained for 1,200 iterations with
a nominal training seed of 42. These are 1,200 fake-score and 239 generator
optimizer steps; the first iteration updates only the fake-score model.
Quality comparisons use matched update budgets, not matched GPU time.

\subsection{Inference Configuration}
\label{app:inference_config}

All reported results are generated using the EMA checkpoints with the same
four-step causal sampling procedure across SF, SGF, and CSF. CSF does not
perform Shortcut Gradient Replay at inference and introduces no additional
inference-time network or forward pass. Videos are generated at
$832\times480$ resolution and 16 fps. We use 21, 243, and 963 latent frames
for the approximately 5-second, 60-second, and 240-second settings,
respectively, which decode to 81, 969, and 3,849 video frames.

For long-video generation, we use a bounded streaming KV cache and keep the
temporal positions within the 21-latent-frame range used during training.
In the chunk-wise setting, the cache contains a three-frame persistent sink,
six recent historical latent frames, and the current three-frame chunk,
for a maximum of 12 latent frames. Temporal positions are assigned using the
top-aligned cache layout. All compared methods use identical cache geometry
and generation seeds. We use a base seed of 42 and assign each generated
sample a deterministic seed according to its prompt/sample ordinal.

\subsection{Evaluation Protocol}
\label{app:evaluation_protocol}

\noindent\textbf{Short-video evaluation.}
For approximately 5-second generation, we use the standard VBench evaluation
pipeline~\citep{VBench}. We generate five samples for each of the 944 unique
prompts in the
official metadata, resulting in 4,720 videos. A frozen manifest explicitly maps
each generated video back to the corresponding VBench metadata entry, resolving
duplicated prompts and filename ambiguity without modifying the underlying
VBench scorers. For consistency with our long-horizon evaluation, we primarily
report the same seven dimensions used for 60-second and 240-second videos.

\noindent\textbf{Long-video evaluation.}
For approximately 60-second and 240-second generation, we use the VBench-Long
implementation in VBench++~\citep{VBenchPP} under a controlled custom-input
protocol. We disable semantic scene splitting and divide every generated video
into fixed 2-second clips. We report seven dimensions: aesthetic quality,
background consistency, dynamic degree, imaging quality, motion smoothness,
subject consistency, and temporal flickering. Temporal flickering is evaluated
over all clips without static filtering. An explicit manifest maps every clip to
its source video, and clip-level scores are aggregated within each source video
before averaging across videos, following the VBench-Long aggregation.
Subject and background consistency retain the VBench-Long cross-clip
comparisons and fuse them with within-clip scores by source video.

The 60-second evaluation uses the official 40 VBench-Long prompts, with one
video generated per prompt. For the 240-second evaluation, we use a fixed set
of 128 prompts sampled without replacement from the 1,003 Movie Gen Video Bench
prompts using a fixed random seed of 42; the sampled prompts are then restored
to their original dataset order. The same frozen prompt sets, generation
settings, clip assignments, and aggregation rules are used for all compared
methods.

\subsection{Initializations and Baseline Reproduction}
\label{app:initializations}

To reduce dependence on a particular initialization, we evaluate three
initialization schemes: teacher forcing (TF init), causal consistency
distillation (Causal CD)~\citep{CausalForcingPP}, and Causal
ODE~\citep{CausalForcing}. These settings cover different causal starting
points and allow us to examine whether the gains of CSF persist across
initialization schemes.

We retrain SF~\citep{SF} and SGF~\citep{SGF} based on their released
implementations. Within each initialization setting, SF, SGF, and CSF are
independently trained from the same initialization checkpoint, using the same
training prompt set, nominal seed, training budget, and inference configuration.
We otherwise preserve the method-specific training procedure of each baseline.
The chunk-wise 60-second and 240-second comparisons constitute our main
results.

\subsection{Shortcut Gradient Replay Implementation}
\label{app:replay_impl}

We further detail the practical implementation of Shortcut Gradient Replay
beyond the algorithmic description in the main paper. During the detached
self-rollout, we store only the states required to reconstruct the local
generation and KV-writing computations, including the sampled exit input and
timestep, generated latent, and corresponding causal cache states. The DMD
output gradients are computed once after the rollout and reused throughout
replay; the recorded stochastic-exit states are also reused rather than
resampled.

Replay is implemented as a single reverse traversal over the autoregressive
chunks. When processing chunk $i$, gradients contributed by later chunks to its
KV state have already been accumulated. The required generation and KV-writing
VJPs are evaluated through separate forward--backward calls, with each temporary
computation graph released immediately after gradient accumulation. Thus, CSF
reconstructs only the required local gradient paths without materializing the
full autoregressive computation graph.

\section{Controlled Gaussian Time-Series Toy Experiment}
\label{sec:toy-concise}

We present a simple toy experiment to motivate our study of cross-step gradient
propagation. This simplified setting provides intuition about its potential
benefits, but the observed behavior need not carry over to real-world video
generation. We compare SF, CSF, and Full BPTT under matched architectures,
initialization, sampling, and neural DMD training, varying the backward paths
through generated history.

\noindent\textbf{Task.}
For independent external inputs $u_j\sim\mathcal{N}(0,1)$, we generate
\begin{equation}
 h_j^{\mathrm{f}}=0.72h_{j-1}^{\mathrm{f}}+u_j,\quad
 h_j^{\mathrm{s}}=0.97h_{j-1}^{\mathrm{s}}+u_j,\quad
 x_j=0.65h_j^{\mathrm{f}}+0.35h_j^{\mathrm{s}}.
\end{equation}
States start at zero, followed by a 96-step burn-in. We retain 2,400 training
sequences of length 128 and standardize observations using statistics from their
first 24 positions. Each rollout starts from four observed positions and their
external inputs. At step $j$, the generator receives the current $u_j$, generated
history, and fresh sampling noise, without future target observations or later
external inputs. Given the observed context and subsequent external inputs, the
target continuation is deterministic, making paired rollout MSE a direct
prediction-error metric.

\noindent\textbf{Training.}
Training proceeds in four stages. We first pretrain a bidirectional
flow-matching teacher for 12,000 updates and freeze it. We initialize the causal
generator with 5,000 updates of teacher-forcing flow matching using ground-truth
history. With the generator fixed, we warm up a bidirectional fake model for
3,000 updates on generated samples. From these shared states, SF, CSF, and Full
BPTT each receive 500 neural DMD generator updates, each preceded by five
fake-model updates; MSE is reserved for evaluation.

The generator shares parameters between velocity prediction and clean KV
writing, using one denoising step per autoregressive position in training and
inference. To reduce gradient variance from random re-noising, we average the
DMD update signal over eight independent re-noising draws of the same generated
sequence. SF detaches generated history; CSF reconnects selected history paths
through shortcut replay; Full BPTT backpropagates through all 20 autoregressive
positions. All methods detach the observed context. We evaluate every endpoint
checkpoint across ten paired seeds with shared minibatch and noise streams.
Table~\ref{tab:toy-concise-config} lists the configuration.

\begin{table}[ht]
\centering\small
\caption{Shared toy configuration. Budgets denote update counts.}
\label{tab:toy-concise-config}
\begin{tabular}{@{}ll@{}}
\toprule
Setting & Value \\
\midrule
Layers / width / heads / history window & 2 / 32 / 4 / 4 \\
Context / training rollout / evaluation rollout & 4 / 20 / 124 positions \\
Generator updates / fake updates per generator update & 500 / 5 \\
Batch size / precision & 64 / float32 \\
Generator / fake learning rate & $5\times10^{-7}$ / $8\times10^{-7}$ \\
AdamW betas / weight decay / gradient clipping & $(0,0.999)$ / 0.01 / 5 \\
DMD noise time & $\mathcal{U}(0.02,0.98)$ \\
\bottomrule
\end{tabular}
\end{table}

\noindent\textbf{Evaluation.}
In \ToyFigureReference, panel (a) tracks validation MSE averaged over positions
21--124 every 25 generator updates on 256 fixed conditions. Panel (b) reports per-position
rollout MSE at update 500 on 8,192 held-out conditions. The larger evaluation
followed a sample-size diagnostic, without retraining or checkpoint selection.
Both use two shared sampling-noise draws. MSE first averages over conditions
and draws, then across ten paired seeds; bands are pointwise 95\% Student-$t$
intervals over seed means, conditional on shared data and initialization.
The unsmoothed curve uses a quadratic y-axis: vertical position is proportional
to MSE squared, with ticks labeled in MSE units; gray shading marks positions
1--20.

\noindent\begin{minipage}{\linewidth}
\textbf{Results.}
As shown in \ToyFigureReference, CSF improves prediction beyond the training
horizon. Over positions 21--124, mean MSE is 0.03088 for SF, 0.02709 for CSF, and
0.02625 for Full BPTT. CSF reduces MSE by 12.27\% relative to SF, improving in
all ten paired seeds, while remaining 3.21\% above Full BPTT.
These results provide intuition within this simplified setting; whether the same
behavior holds in real-world video generation requires separate evaluation.
\end{minipage}

\begin{table*}[t]
\centering
\caption{
Training efficiency and peak single-GPU memory (GiB). Device denotes sampled
NVML usage; Allocated and Reserved denote PyTorch peaks. Time is normalized to
SF separately within each setting. Full BPTT's
frame-wise OOM has no completed-run peaks or time.
}
\label{tab:training_efficiency_memory}
\small
\setlength{\tabcolsep}{7pt}
\begin{tabular}{llcccc}
\toprule
Setting & Method & Device & Allocated & Reserved & Time ($\times$ SF) \\
\midrule
\multirow{3}{*}{Chunk-wise}
& SF        & 103.65 & 70.64 & 98.46 & 1.00$\times$ \\
& Full BPTT & 148.65 & 118.70 & 143.45 & 1.15$\times$ \\
& CSF       & \textbf{88.46} & 59.02 & 83.26 & 1.27$\times$ \\
\midrule
\multirow{3}{*}{Frame-wise}
& SF        & 184.48 & 149.44 & 179.28 & 1.00$\times$ \\
& Full BPTT & OOM    & -- & -- & -- \\
& CSF       & \textbf{88.42} & 59.01 & 83.22 & 1.41$\times$ \\
\bottomrule
\end{tabular}
\end{table*}

\begin{table*}[t]
\centering
\caption{Full 16-dimensional VBench results for 5-second chunk-wise generation
under three causal initializations. Avg.(6) is the unweighted mean of Aesthetic Quality, Background Consistency,
Imaging Quality, Motion Smoothness, Subject Consistency, and Flickering,
excluding Dynamic Degree. Except for Dynamic Degree, best and second-best results within each
initialization are shown in \textbf{bold} and \underline{underlined},
respectively.}
\label{tab:chunkwise_5s}
\small
\setlength{\tabcolsep}{3.0pt}
\renewcommand{\arraystretch}{0.95}

\begin{tabular}{lccc ccc ccc}
\toprule
& \multicolumn{3}{c}{Causal CD} & \multicolumn{3}{c}{TF} & \multicolumn{3}{c}{Causal ODE} \\
\cmidrule(lr){2-4}
\cmidrule(lr){5-7}
\cmidrule(lr){8-10}

Metric
& SF & SGF & \textbf{CSF (Ours)}
& SF & SGF & \textbf{CSF (Ours)}
& SF & SGF & \textbf{CSF (Ours)} \\
\midrule

Aesthetic
& \underline{0.6430} & 0.6393 & \textbf{0.6515} & \underline{0.6593} & 0.6527 & \textbf{0.6726} & \textbf{0.6566} & \underline{0.6496} & 0.6490 \\

Appear.
& \underline{0.1890} & 0.1871 & \textbf{0.1902} & \underline{0.1895} & \textbf{0.1916} & 0.1886 & 0.1939 & \underline{0.1951} & \textbf{0.1963} \\

Background
& \underline{0.9554} & 0.9551 & \textbf{0.9591} & \underline{0.9457} & 0.9349 & \textbf{0.9739} & \textbf{0.9679} & 0.9327 & \underline{0.9612} \\

Color
& \textbf{0.9327} & \underline{0.9099} & 0.8963 & 0.8876 & \textbf{0.9021} & \underline{0.8987} & \underline{0.9034} & 0.8908 & \textbf{0.9090} \\

Action
& 0.7180 & \textbf{0.7500} & \underline{0.7300} & 0.7460 & \underline{0.7540} & \textbf{0.7620} & 0.7520 & \underline{0.7640} & \textbf{0.7660} \\

Imaging
& \textbf{0.7130} & 0.7088 & \underline{0.7117} & 0.7053 & \textbf{0.7184} & \underline{0.7116} & \underline{0.7156} & 0.7128 & \textbf{0.7196} \\

Motion
& \textbf{0.9878} & 0.9821 & \underline{0.9853} & \underline{0.9863} & 0.9710 & \textbf{0.9886} & 0.9879 & \underline{0.9881} & \textbf{0.9900} \\

Multi-Obj
& \textbf{0.6466} & 0.5590 & \underline{0.6352} & \textbf{0.6991} & 0.6591 & \underline{0.6630} & \textbf{0.6951} & \underline{0.6899} & 0.6309 \\

Object
& \textbf{0.8062} & 0.7962 & \underline{0.8027} & \underline{0.8392} & 0.8201 & \textbf{0.8593} & \textbf{0.8418} & \underline{0.8400} & 0.8203 \\

Overall
& \underline{0.2382} & 0.2341 & \textbf{0.2398} & \textbf{0.2455} & 0.2377 & \underline{0.2413} & \underline{0.2463} & \textbf{0.2474} & 0.2388 \\

Scene
& \underline{0.2738} & 0.2320 & \textbf{0.2749} & \textbf{0.3126} & 0.2631 & \underline{0.3097} & \textbf{0.3060} & \underline{0.3047} & 0.2728 \\

Spatial
& \textbf{0.8005} & 0.7234 & \underline{0.7767} & \textbf{0.8391} & 0.7917 & \underline{0.8227} & \underline{0.8013} & \textbf{0.8185} & 0.7683 \\

Subject
& \textbf{0.9748} & 0.9640 & \underline{0.9703} & \underline{0.9645} & 0.9534 & \textbf{0.9782} & \underline{0.9765} & 0.9725 & \textbf{0.9772} \\

Flickering
& 0.9900 & 0.9877 & 0.9900 & \underline{0.9910} & 0.9540 & \textbf{0.9954} & 0.9901 & \underline{0.9909} & \textbf{0.9931} \\

Temp.Style
& \textbf{0.2359} & 0.2328 & \underline{0.2347} & 0.2315 & \underline{0.2318} & \textbf{0.2319} & \textbf{0.2419} & \underline{0.2385} & 0.2332 \\

Dynamic
& 0.4944 & 0.7194 & 0.6028 & 0.6250 & 0.9583 & 0.5111 & 0.4389 & 0.5667 & 0.4944 \\
\midrule

Avg.(6)
& \underline{0.8773} & 0.8728 & \textbf{0.8780} & \underline{0.8753} & 0.8641 & \textbf{0.8867} & \textbf{0.8824} & 0.8744 & \underline{0.8817} \\
\bottomrule
\end{tabular}
\end{table*}

\begin{figure*}[t]
  \centering
  \includegraphics[width=0.92\textwidth]{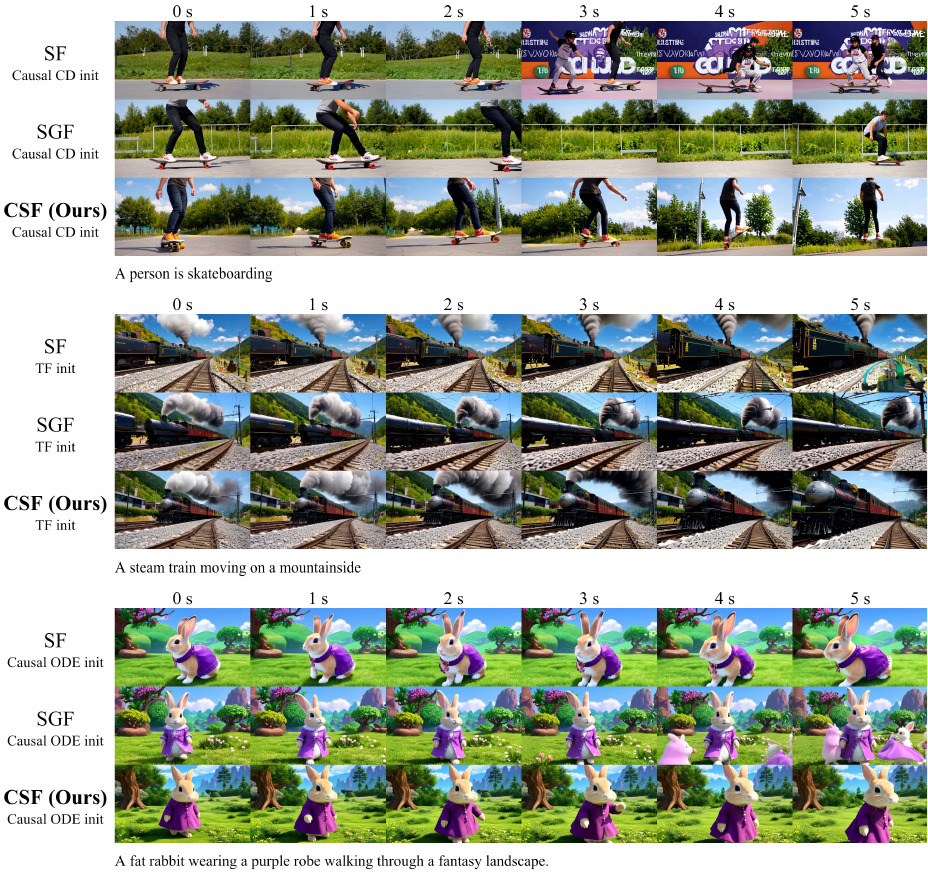}
  \caption{Additional qualitative comparison of 5-second chunk-wise generation.
    We compare SF, SGF, and CSF under Causal CD, TF, and Causal ODE
    initialization using uniformly spaced snapshots.}
  \label{fig:supp_5s}
\end{figure*}

\section{Further Experiments}
\label{app:further_experiments}

\subsection{Training Efficiency and Memory}
\label{app:training_efficiency}

We profile SF, replay-based full backpropagation
through the autoregressive rollout (Full BPTT), and CSF. Each uses eight
processes (one GPU and batch size one per process), BF16, hybrid-full FSDP,
gradient checkpointing, and matched allocator and cache-release settings.
The chunk-wise test follows our main rollout configuration; the frame-wise
stress test uses 21 autoregressive latent steps with a nine-latent local
history.

For a controlled comparison, all methods execute the complete four-step
denoising rollout. For each video on each GPU, we independently sample one
denoising step as the gradient-enabled exit step and use the same exit step for
all autoregressive units in that video; the remaining denoising steps do not
contribute generator gradients. For Full BPTT, we first record the complete
rollout without autograd, and then replay the selected denoising step together
with the KV-writing transition for each autoregressive unit under autograd,
while retaining the full historical computation graph across the rollout. This
allows the gradient signals from later autoregressive units to propagate
through the connected historical KV states in a single backward pass.

For completed runs, peak device memory is the maximum sampled device-wide
NVML \texttt{memory.used} on any GPU across all 50 iterations; allocated and
reserved are maxima of per-step PyTorch peaks.
Time is the slowest rank's elapsed time from the end of iteration 5 through
the end of iteration 50, divided by nine. Each five-iteration unit includes
one generator and five fake-score updates. Table~\ref{tab:training_efficiency_memory}
reports the resulting single-GPU memory and per-five-iteration time,
normalized to SF within each setting.

In the chunk-wise setting, replay-based Full BPTT reaches 1.43$\times$ the
SF peak device memory. CSF uses 0.85$\times$ the measured SF peak at
1.27$\times$ the time. In the frame-wise stress test, Full BPTT runs out of
memory on its first generator update; CSF completes 50 iterations at
0.48$\times$ the measured SF peak and 1.41$\times$ the time. The reported memory and time ratios are specific to the standardized full-four-step profiling configuration.

Full BPTT reconstructs and retains the connected computation graph across the
autoregressive history. CSF instead replays selected paths and releases each
local graph after backward. CSF's lower measured peak memory is consistent
with its replay schedule and early release of local graphs. It does not imply
that adding cross-chunk gradients intrinsically reduces memory.

In a separate chunk-wise diagnostic, direct Full BPTT with autograd enabled
throughout the forward rollout reached 192.50\,GiB. We therefore use
replay-based Full BPTT as the more memory-efficient control in
Table~\ref{tab:training_efficiency_memory}: it shares the detached four-step
rollout and selected-exit reconstruction with the other methods, yet retains
all reconstructed historical connections until a single backward pass. The
explicit replay cost is included in the reported
time; gradient checkpointing is applied separately to all methods.

\begin{figure*}[t]
  \centering
  \includegraphics[width=0.92\textwidth]{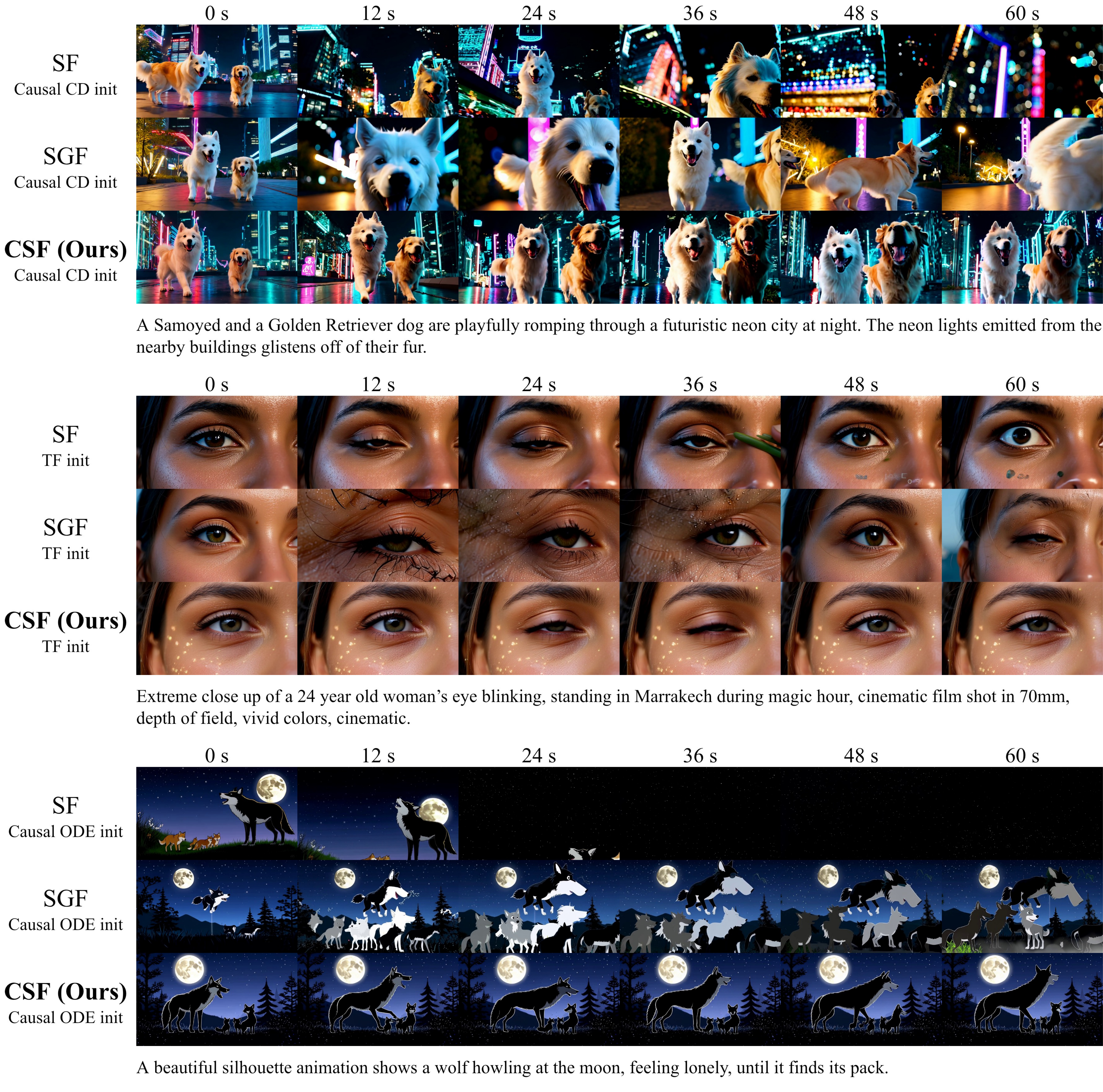}
  \caption{Additional qualitative comparison of 60-second chunk-wise
    generation. We compare SF, SGF, and CSF under Causal CD, TF, and Causal
    ODE initialization using uniformly spaced snapshots throughout the
    rollout.}
  \label{fig:supp_60s}
\end{figure*}

\subsection{Short-Horizon Chunk-wise Evaluation}
\label{app}

To complement the long-horizon evaluations in the main paper, we further
provide the full 16-dimensional VBench results
(Table~\ref{tab:chunkwise_5s}) for approximately 5-second chunk-wise
generation under the same three initialization settings: Causal CD, TF, and
Causal ODE. We additionally report Avg.(6), defined as the unweighted
average of Aesthetic Quality, Background Consistency, Imaging Quality,
Motion Smoothness, Subject Consistency, and Flickering.

Overall, SF, SGF, and CSF remain broadly comparable at this short horizon.
CSF achieves the highest Avg.(6) under Causal CD and TF. Under Causal ODE,
CSF remains closely matched with SF in Avg.(6). The individual VBench
dimensions exhibit mixed rankings across methods rather than a uniform
advantage for any single method. The six-metric average remains broadly comparable at short horizons,
while individual semantic and compositional dimensions show
setting-dependent trade-offs.

\begin{figure*}[t]
  \centering
  \includegraphics[width=0.92\textwidth]{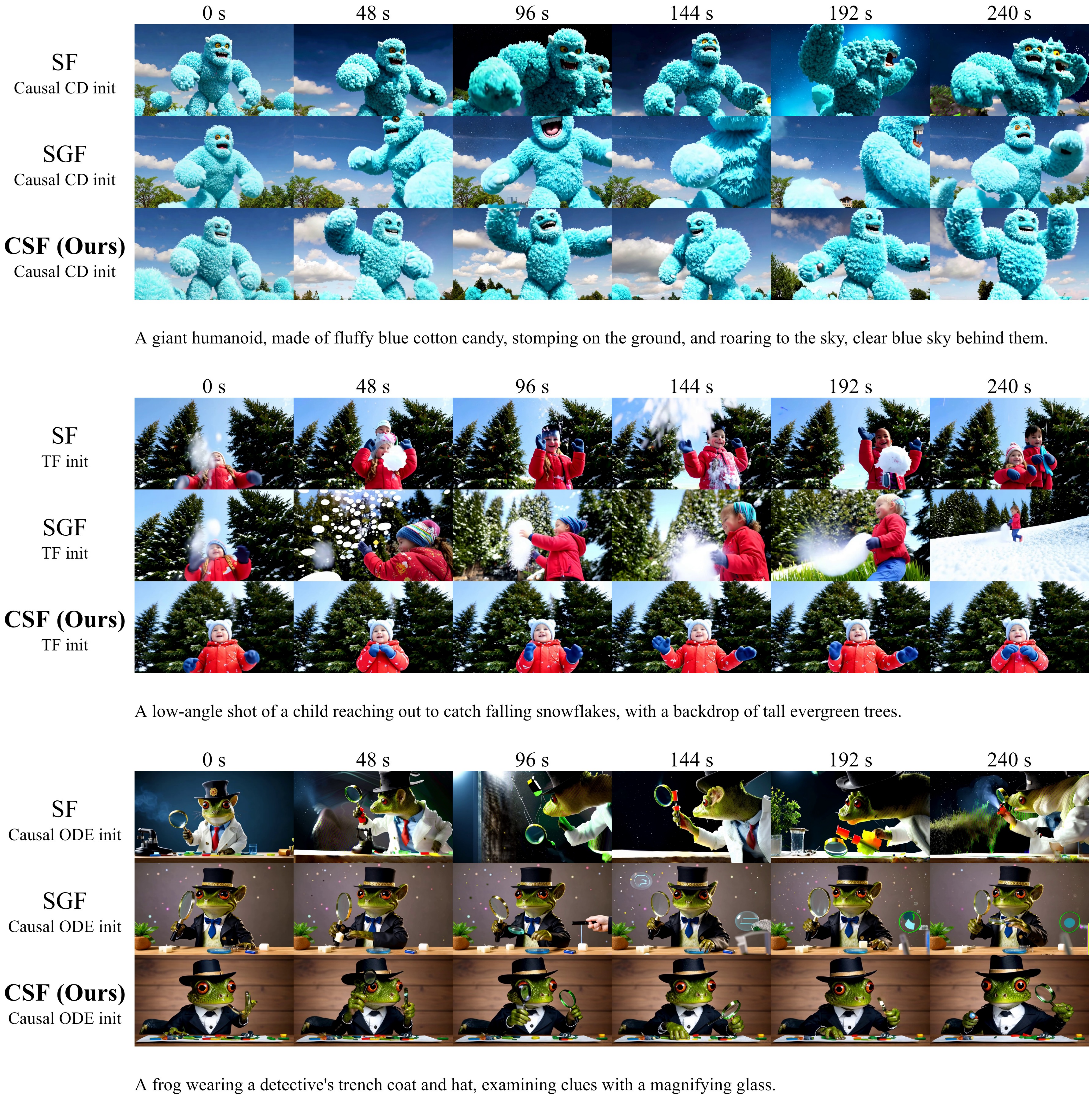}
  \caption{Additional qualitative comparison of 240-second chunk-wise
    generation. We compare SF, SGF, and CSF under Causal CD, TF, and Causal
    ODE initialization using uniformly spaced snapshots over the four-minute
    rollout.}
  \label{fig:supp_240s}
\end{figure*}

\subsection{Additional Chunk-wise Qualitative Results}
\label{app:additional_chunkwise_qualitative}

We provide additional qualitative comparisons for chunk-wise autoregressive
generation at 5, 60, and 240 seconds. These examples complement the
quantitative results in the main paper and the short-horizon evaluation
above, covering both standard short-video generation and substantially
longer autoregressive rollouts. We use the same initialization settings and
evaluation prompts as in the corresponding quantitative experiments.

Figures~\ref{fig:supp_5s}--\ref{fig:supp_240s} present additional
comparisons among SF, SGF, and CSF. At 5 seconds, the three methods
generally produce comparable visual quality, with only modest variations in
subject appearance and local scene details. As the rollout extends to 60
seconds, longer-term drift becomes easier to observe: the appearance and
composition of the subjects may gradually change, and in some cases the
generated content departs more noticeably from the earlier part of the
sequence. At 240 seconds, these effects become more apparent in examples
involving articulated characters, human subjects, and animals, where
subject identity, scale, and surrounding scene structure can evolve over
time. In the examples shown, CSF tends to preserve the defining visual
characteristics of the subject and the overall composition more steadily
across the rollout. These qualitative results are intended to complement
the long-horizon quantitative evaluation in the main paper.

\subsection{Frame-wise Experiments}
\label{app:framewise}

To further evaluate CSF beyond chunk-wise autoregressive generation, we
conduct additional frame-wise experiments. All methods are trained and
evaluated with a matched nine-latent history window, consisting of three
sink and six recent latents, comparable in scale to the chunk-wise setting.
Training lasts 1,500 iterations, corresponding to 1,500 fake-score and 299
generator optimizer steps. The results therefore reflect this matched-cache
setting rather than each baseline's native configuration. All other settings
follow the corresponding chunk-wise experiments unless otherwise specified.

Table~\ref{tab:framewise_long_horizon} reports the 60- and 240-second results
under Causal CD and TF initialization. Under Causal CD, CSF achieves the
highest Avg.(6) at both horizons. Under TF, CSF slightly improves over SF at
60 seconds and remains closely matched at 240 seconds; the paired bootstrap
intervals include zero at both horizons (Appendix~\ref{app:paired_bootstrap}).
Overall, these results support the applicability of CSF to finer-grained
frame-wise autoregression, while indicating that the gains depend on the
initialization setting.

\begin{figure*}[t]
  \centering
  \includegraphics[width=0.90\textwidth]{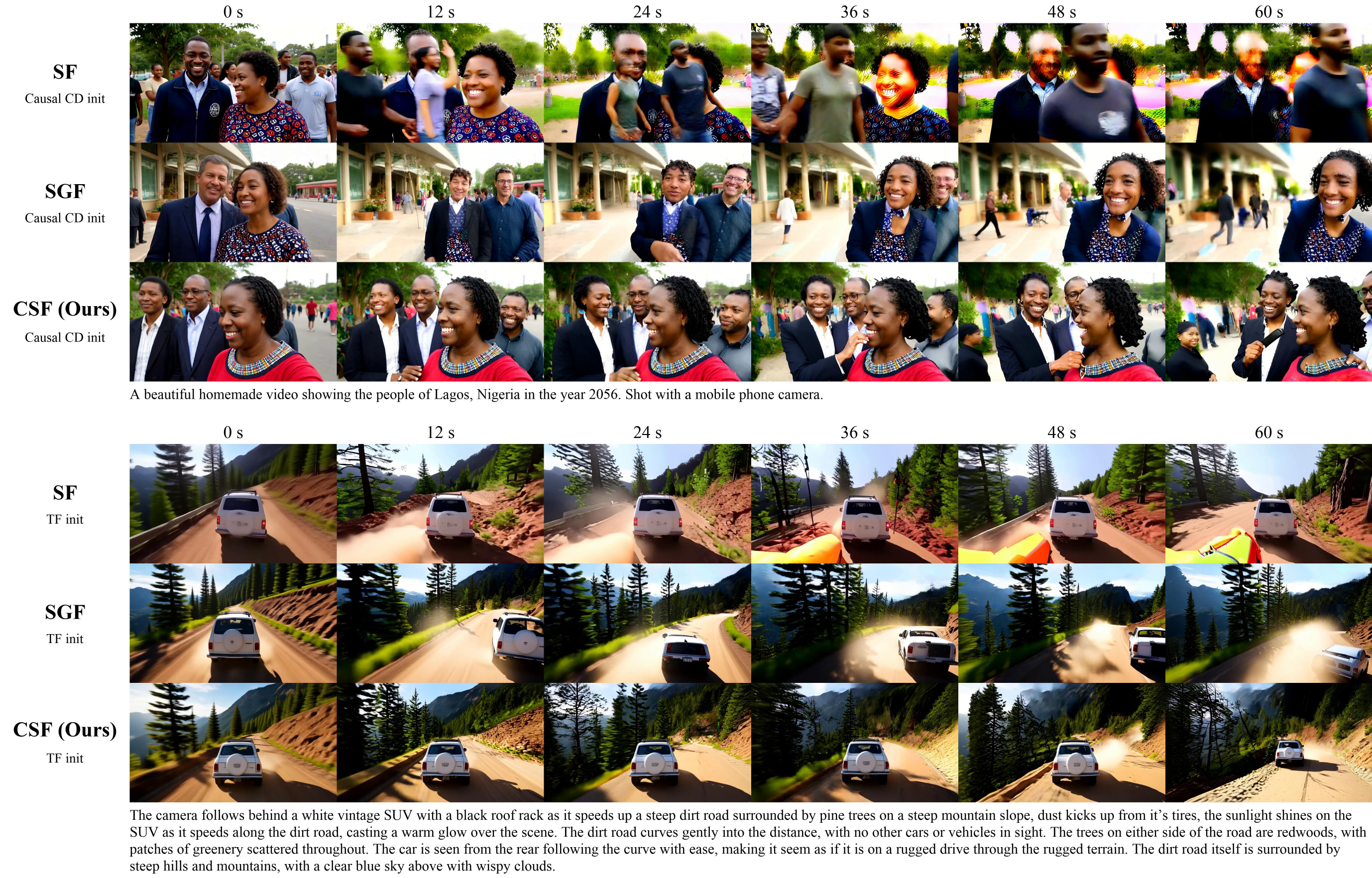}
  \caption{Additional qualitative comparison of 60-second frame-wise
    generation. We compare SF, SGF, and CSF under Causal CD and TF
    initialization using uniformly spaced snapshots throughout the
    rollout.}
  \label{fig:framewise_qual_60s}
\end{figure*}

\begin{figure*}[t]
  \centering
  \includegraphics[width=0.90\textwidth]{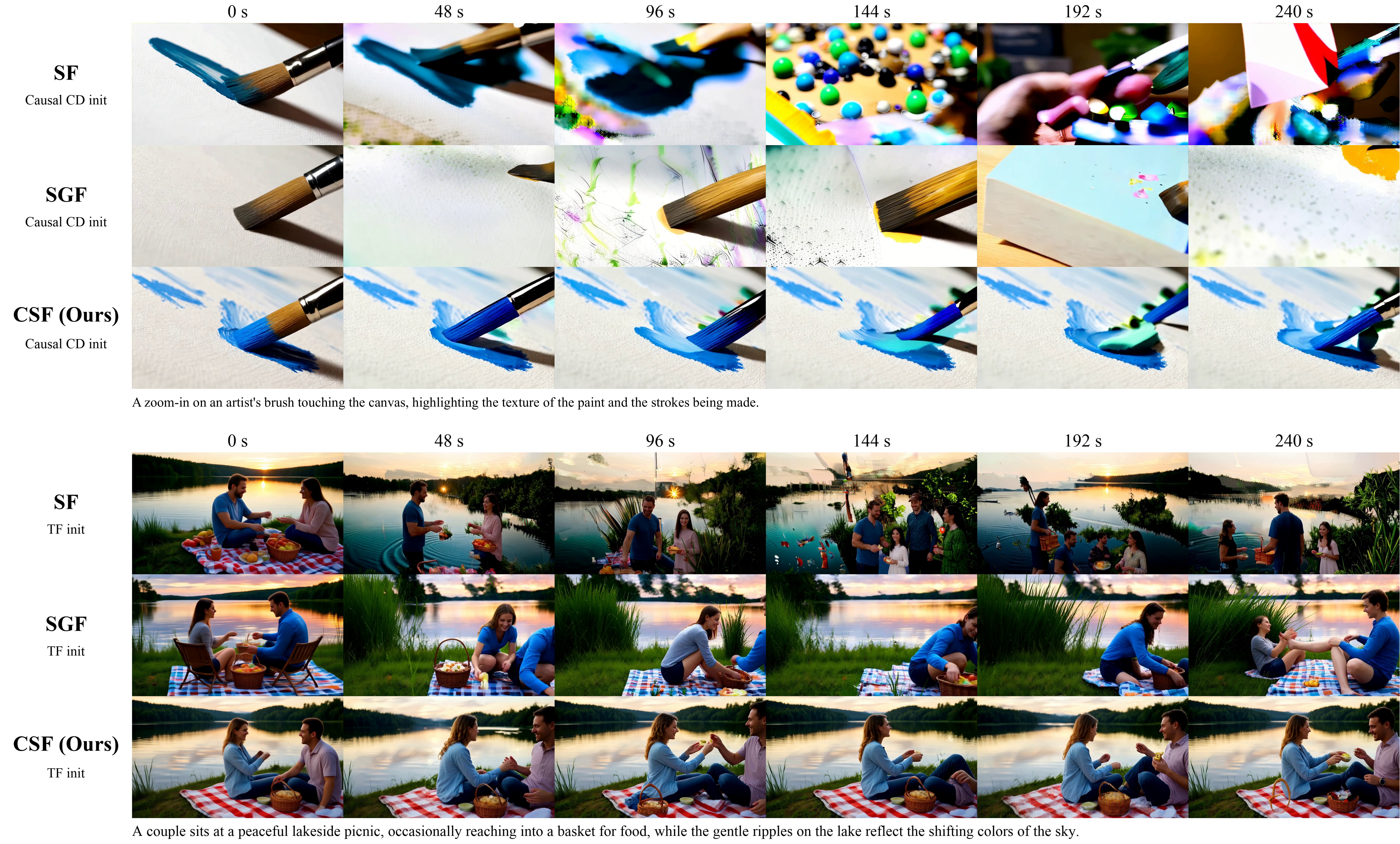}
  \caption{Additional qualitative comparison of 240-second frame-wise
    generation. We compare SF, SGF, and CSF under Causal CD and TF
    initialization using uniformly spaced snapshots over the four-minute
    rollout.}
  \label{fig:framewise_qual_240s}
\end{figure*}

Figures~\ref{fig:framewise_qual_60s} and~\ref{fig:framewise_qual_240s} provide
additional qualitative comparisons at 60 and 240 seconds. As the autoregressive
rollout becomes longer, variations in subject appearance, composition, and
scene content become increasingly visible in some examples. In the examples
shown, CSF tends to maintain the key visual characteristics of the subject
and the overall scene structure more steadily over time, complementing the
quantitative results above.

\begin{table*}[t]
\centering
\caption{Frame-wise long-horizon quantitative comparison at 60 and 240 seconds.
We compare SF, SGF, and CSF under Causal CD and TF initialization using the
same evaluation protocol as the chunk-wise experiments. Avg.(6) and highlighting
conventions follow Table~\ref{tab:chunkwise_5s}.}
\label{tab:framewise_long_horizon}
\small
\setlength{\tabcolsep}{2.5pt}
\renewcommand{\arraystretch}{0.95}

\begin{tabular}{lllcccccccc}
\toprule
Horizon & Init. & Method & Aes.$\uparrow$ & Back.$\uparrow$ & Imag.$\uparrow$ & Mot.$\uparrow$ & Subj.$\uparrow$ & Flick.$\uparrow$ & Dyn. & Avg.(6)$\uparrow$ \\
\midrule

 &  & SF & \underline{0.5159} & 0.9453 & 0.6366 & 0.9614 & 0.9366 & 0.9376 & 0.9411 & 0.8222 \\
 & Causal CD & SGF & 0.5103 & \underline{0.9517} & \underline{0.6408} & \underline{0.9834} & \underline{0.9475} & \underline{0.9659} & 0.7234 & \underline{0.8333} \\
60 s &  & \textbf{CSF (Ours)} & \textbf{0.5170} & \textbf{0.9586} & \textbf{0.6805} & \textbf{0.9861} & \textbf{0.9671} & \textbf{0.9699} & 0.6589 & \textbf{0.8465} \\
\cmidrule(lr){2-11}
 &  & SF & \textbf{0.6013} & \underline{0.9601} & \textbf{0.7380} & \underline{0.9777} & \underline{0.9705} & \underline{0.9629} & 0.7927 & \underline{0.8684} \\
 & TF & SGF & 0.5866 & 0.9533 & 0.7234 & 0.9773 & 0.9632 & 0.9478 & 0.8782 & 0.8586 \\
 &  & \textbf{CSF (Ours)} & \underline{0.5871} & \textbf{0.9634} & \underline{0.7328} & \textbf{0.9869} & \textbf{0.9763} & \textbf{0.9723} & 0.6516 & \textbf{0.8698} \\
\midrule

 &  & SF & \textbf{0.4833} & 0.9437 & \textbf{0.6056} & 0.9711 & 0.9318 & 0.9470 & 0.9461 & \underline{0.8137} \\
 & Causal CD & SGF & 0.4406 & \underline{0.9484} & 0.5505 & \underline{0.9856} & \underline{0.9404} & \underline{0.9712} & 0.6890 & 0.8061 \\
240 s &  & \textbf{CSF (Ours)} & \underline{0.4490} & \textbf{0.9535} & \underline{0.5960} & \textbf{0.9879} & \textbf{0.9566} & \textbf{0.9737} & 0.6047 & \textbf{0.8194} \\
\cmidrule(lr){2-11}
 &  & SF & \textbf{0.5868} & \underline{0.9583} & \textbf{0.7228} & \underline{0.9792} & \underline{0.9696} & \underline{0.9658} & 0.7771 & \textbf{0.8638} \\
 & TF & SGF & 0.5586 & 0.9533 & 0.7000 & 0.9771 & 0.9620 & 0.9512 & 0.9162 & 0.8504 \\
 &  & \textbf{CSF (Ours)} & \underline{0.5701} & \textbf{0.9632} & \underline{0.7116} & \textbf{0.9853} & \textbf{0.9744} & \textbf{0.9707} & 0.7357 & \underline{0.8626} \\
\bottomrule
\end{tabular}
\end{table*}

\subsection{Paired Bootstrap Analysis}
\label{app:paired_bootstrap}

We assess evaluation-sample uncertainty in Avg.(6) for the chunk-wise and
frame-wise long-video settings reported above. The resampling unit is a matched
prompt and its original full-length video; clip-level scores are first
aggregated within that video, and its six dimension scores are averaged to
obtain Avg.(6). Within each initialization and duration, we
resample the same video identities for CSF and each baseline with replacement
20,000 times ($N=40$ at 60 seconds and $N=128$ at 240 seconds). We report
CSF-minus-baseline mean differences with 95\% percentile intervals from the
2.5th and 97.5th bootstrap percentiles. These comparison-wise intervals
describe evaluation-sample uncertainty conditional on fixed checkpoints, not
variation across training seeds.

\begin{table}[!ht]
\centering
\caption{Paired bootstrap differences in Avg.(6) for the reported long-video
settings. Each entry is the difference computed from unrounded per-video scores
with its 95\% percentile interval, multiplied by 100 (score points). Positive
values favor CSF; intervals are not adjusted for multiple comparisons.}
\label{tab:paired_bootstrap_avg6}
\footnotesize
\setlength{\tabcolsep}{6pt}
\begin{tabular}{llrll}
\toprule
Initialization & Duration & $N$ & CSF$-$SF & CSF$-$SGF \\
\midrule
\multicolumn{5}{l}{\textbf{(a) Chunk-wise}} \\
Causal CD & 60 s  & 40  & $+0.519\;[-0.228,+1.250]$ & $+0.261\;[-0.350,+0.826]$ \\
          & 240 s & 128 & $+1.000\;[+0.657,+1.328]$ & $+1.144\;[+0.870,+1.426]$ \\
TF        & 60 s  & 40  & $+1.326\;[+0.798,+1.861]$ & $+1.977\;[+1.424,+2.540]$ \\
          & 240 s & 128 & $+1.560\;[+1.287,+1.829]$ & $+2.065\;[+1.739,+2.395]$ \\
Causal ODE & 60 s  & 40  & $+1.813\;[+1.278,+2.390]$ & $+0.068\;[-0.385,+0.528]$ \\
           & 240 s & 128 & $+1.000\;[+0.688,+1.323]$ & $+0.186\;[-0.128,+0.493]$ \\
\midrule
\multicolumn{5}{l}{\textbf{(b) Frame-wise}} \\
Causal CD & 60 s  & 40  & $+2.433\;[+1.247,+3.659]$ & $+1.330\;[+0.534,+2.126]$ \\
          & 240 s & 128 & $+0.570\;[+0.022,+1.116]$ & $+1.332\;[+0.846,+1.820]$ \\
TF        & 60 s  & 40  & $+0.138\;[-0.300,+0.595]$ & $+1.119\;[+0.673,+1.609]$ \\
          & 240 s & 128 & $-0.119\;[-0.403,+0.164]$ & $+1.219\;[+0.944,+1.498]$ \\
\bottomrule
\end{tabular}
\end{table}

\subsection{Exploring Gradient Conflict}
\label{app:gradient_conflict}

CSF combines the direct DMD gradient $g_{\mathrm{DMD}}$ with cross-chunk
gradient feedback $g_{\mathrm{cross}}$ propagated from future blocks. In
practice, $g_{\mathrm{cross}}$ can contain components that are negatively
aligned with $g_{\mathrm{DMD}}$. To examine whether such conflicts should be
explicitly suppressed, we consider a simple conflict-aware variant that
projects out the opposing component whenever
$g_{\mathrm{cross}}^\top g_{\mathrm{DMD}} < 0$:
\begin{equation}
g_{\mathrm{cross}}^{\mathrm{proj}}
=
g_{\mathrm{cross}}
-
\frac{
\min(g_{\mathrm{cross}}^\top g_{\mathrm{DMD}},0)
}{
\|g_{\mathrm{DMD}}\|_2^2+\epsilon
}
g_{\mathrm{DMD}}.
\end{equation}
This variant is used only for analysis and is not part of the default CSF
formulation.

Table~\ref{tab:grad_projection} compares the default CSF with this projected
variant at 5, 60, and 240 seconds. Removing the negatively aligned component
consistently increases Dynamic Degree, while Avg.(6) decreases slightly across
all three horizons. This suggests that gradient components opposing the direct
DMD objective are not necessarily purely detrimental. One possible
interpretation is that part of the negative-direction feedback from
$g_{\mathrm{cross}}$ acts as a damping signal, encouraging more conservative
temporal evolution and thereby reducing motion dynamics. Removing this
component shifts the generation toward stronger temporal activity, but does
not yield a consistent improvement in visual quality or long-term consistency.
We therefore retain the original, unprojected cross-chunk gradient feedback in
the default CSF formulation.

\begin{table*}[t]
\centering
\caption{Effect of conflict-aware gradient projection on CSF under Causal CD
at 5, 60, and 240 seconds. GradProj removes the component of
$g_{\mathrm{cross}}$ that is negatively aligned with $g_{\mathrm{DMD}}$.}
\label{tab:grad_projection}
\small
\setlength{\tabcolsep}{3.0pt}
\renewcommand{\arraystretch}{0.95}

\begin{tabular}{llcccccccc}
\toprule
Duration & Method & Aes.$\uparrow$ & Back.$\uparrow$ & Imag.$\uparrow$ & Mot.$\uparrow$ & Subj.$\uparrow$ & Flick.$\uparrow$ & Dyn. & Avg.(6)$\uparrow$ \\
\midrule

5 s & \textbf{CSF} & \textbf{0.6515} & \textbf{0.9591} & \textbf{0.7117} & \textbf{0.9853} & \textbf{0.9703} & \textbf{0.9900} & 0.6028 & \textbf{0.8780} \\
 & + GradProj & 0.6365 & 0.9471 & 0.7075 & 0.9822 & 0.9606 & 0.9775 & 0.7083 & 0.8686 \\
\midrule

60 s & \textbf{CSF} & \textbf{0.5659} & \textbf{0.9597} & 0.7085 & \textbf{0.9848} & \textbf{0.9707} & \textbf{0.9637} & 0.5976 & \textbf{0.8589} \\
 & + GradProj & 0.5436 & 0.9543 & \textbf{0.7175} & 0.9835 & 0.9617 & 0.9614 & 0.7605 & 0.8537 \\
\midrule

240 s & \textbf{CSF} & \textbf{0.5449} & \textbf{0.9563} & \textbf{0.6938} & \textbf{0.9850} & \textbf{0.9666} & \textbf{0.9656} & 0.6899 & \textbf{0.8520} \\
 & + GradProj & 0.5254 & 0.9527 & 0.6795 & 0.9832 & 0.9598 & 0.9630 & 0.7997 & 0.8439 \\
\bottomrule
\end{tabular}
\end{table*}

\section{Discussion and Limitations}
\label{app:limitations}

CSF restores future-to-history gradient feedback without retaining the full
autoregressive computation graph, but it does not aim to reproduce exact
full-sequence backpropagation. In particular, Shortcut Gradient Replay
selectively reconstructs paths to historical chunks, whereas Full BPTT
connects the replayed computation across the autoregressive history. Its
additional paths may give early chunks too much future feedback relative to
direct DMD supervision. This suggests that the goal is not necessarily to
propagate gradients through as much history as possible, but rather to provide
useful future feedback while preserving a sufficiently strong local learning
signal. This controlled comparison does not establish that full-history
backpropagation is generally worse; further study is needed to understand
when it helps. How to determine which historical states should receive gradient
feedback, and from which future states, remains an open question.

CSF trades memory consumption for additional computation through replay.
Although the resulting training overhead remains practical in our
post-training setting, replay becomes more expensive as the number of replayed historical
states or writer dependencies grows. This may become more
relevant for larger models or substantially longer autoregressive sequences.
A natural direction is therefore to make historical gradient propagation more
selective, for example through selective replay or sparse historical gradient
assignment, so that gradients are propagated only through historical states
that are most relevant to future generation. Such mechanisms could potentially
retain the benefit of future-to-history feedback while reducing unnecessary
recomputation and gradient interference.

Our experiments also suggest a possible interaction between long-term
consistency and temporal dynamics. In particular, removing components of the
cross-chunk gradient that oppose the direct DMD gradient increases Dynamic
Degree, while slightly reducing the aggregated quality and consistency
metrics. This indicates that some future-to-history feedback may encourage
more conservative temporal evolution rather than simply acting as harmful
gradient conflict. Although CSF does not always achieve the highest Dynamic
Degree in the current unconstrained generation setting, this trade-off may
become less limiting in controllable generation or world-model settings, where
motion and scene evolution are driven more explicitly by control signals,
actions, or other external conditions. In such settings, CSF may be
particularly useful for further reducing autoregressive error accumulation and
preserving visual or state consistency over long horizons, while the desired
dynamics are governed by the control input itself. Better understanding how
future-to-history feedback should interact with controllable dynamics is
therefore an interesting direction for future work.